%% file: main.tex
\documentclass[11pt,a4paper]{article}
\usepackage[hyperref]{acl2020}
\usepackage{times}
\usepackage{latexsym}
\renewcommand{\UrlFont}{\ttfamily\small}
\usepackage{blindtext}
\usepackage{etoc}

\usepackage{microtype}

\aclfinalcopy 

\usepackage{graphicx}
\usepackage{enumitem}

\usepackage{xspace,mfirstuc,tabulary}
\usepackage{latexsym}
\usepackage{amsthm,amsmath,amsfonts,amssymb}
\usepackage[noend]{algpseudocode}
\usepackage[ruled,vlined]{algorithm2e}
\usepackage[]{mcode}
\usepackage{comment}
\usepackage{multirow}
\usepackage{bigstrut}
\usepackage{url}
\usepackage{subfigure}
\usepackage{color, colortbl} 
\usepackage{xcolor}
\usepackage{booktabs}
\usepackage{sidecap}
\usepackage{verbatim} 
\usepackage{pifont}
\usepackage{eqparbox}
\usepackage{arydshln} 
\usepackage[linewidth=1pt]{mdframed}
\usepackage{framed}
\usepackage{epigraph}
\usepackage{lipsum}
\usepackage[utf8]{inputenc} 
\usepackage[makeroom]{cancel}
\usepackage{tikz}
\usepackage{cases}
\usepackage[colorinlistoftodos]{todonotes}

\newtheorem{theo}{Theorem}
\newenvironment{theorem}
  {\begin{mdframed}\begin{theo}}
  {\end{theo}\end{mdframed}}

\newtheorem{remark1}[theo]{Remark}
\newenvironment{remark}
  {\begin{mdframed}\begin{remark1}}
  {\end{remark1}\end{mdframed}}

\newtheorem{def1}[theo]{Definition}
\newenvironment{definition}
  {\begin{mdframed}\begin{def1}}
  {\end{def1}\end{mdframed}}

\newtheorem{lem1}[theo]{Lemma}
\newenvironment{lemma}
  {\begin{mdframed}\begin{lem1}}
  {\end{lem1}\end{mdframed}}

\newtheorem{prop1}[theo]{Proposition}
\newenvironment{proposition}
  {\begin{mdframed}\begin{prop1}}
  {\end{prop1}\end{mdframed}}

\newtheorem{cor1}[theo]{Corollary}
\newenvironment{corollary}
  {\begin{mdframed}\begin{cor1}}
  {\end{cor1}\end{mdframed}}

\newcolumntype{L}[1]{>{\raggedright\let\newline\\\arraybackslash\hspace{0pt}}m{#1}}
\newcolumntype{C}[1]{>{\centering\let\newline\\\arraybackslash\hspace{0pt}}m{#1}}

\newcommand{\ignore}[1]{}

\definecolor{purple}{rgb}{0.5,0,1}
\definecolor{dcyan}{rgb}{0.2,0.6,0.5}
\definecolor{darkgreen}{rgb}{0,200,0}
\definecolor{light-gray}{gray}{0.95} 
\definecolor{darkgreen}{RGB}{0,140,0}
\definecolor{darkred}{RGB}{200,0,0}
\definecolor{lightgreen}{RGB}{231,255,219}
\definecolor{lightred}{RGB}{252,231,234}
\definecolor{lightyellow}{RGB}{250,253,191}
\definecolor{DarkRed}{RGB}{130,25,0}

\newcommand{\daniel}[1]{{\textcolor{blue}{[DK: {#1}]}}}

\newcommand{\changed}[1]{{\color{black}{#1}}}

\newcommand{\norm}[1]{\| #1 \|}
\newcommand{\dist}{\mathrm{dist}}
\newcommand{\pathk}[1]{\overset{#1}{\leftrightsquigarrow}}
\newcommand{\npathk}[1]{\overset{#1}{\cancel{\leftrightsquigarrow}}}
\newcommand{\prob}[1]{\mathbb{P}\left[#1 \right]}
\newcommand{\probTwo}[2]{\mathbb{P}^{(#1)}\left[#2 \right]}
\newcommand{\eone}{\pathk{d}}
\newcommand{\etwo}{\npathk{}}

\newcommand{\meaningspace}{\colorbox{lightred}{meaning space}\xspace}
\newcommand{\linguisticspace}{\colorbox{lightgreen}{linguistic space}\xspace}
\newcommand{\utterance}{\colorbox{lightyellow}{utterence}\xspace}
\newcommand{\meaninggraph}{\colorbox{lightred}{meaning graph}\xspace}
\newcommand{\linguisticgraph}{\colorbox{lightgreen}{linguistic graph}\xspace}

\newcommand*\circled[1]{\tikz[baseline=(char.base)]{
             \node[shape=circle,draw,inner sep=0.2pt] (char) {\tiny #1};}}

\newcommand{\namecite}[1]{\citeauthor{#1}~\shortcite{#1}}

\title{On the Capabilities and Limitations of \daniel{Multi-hop} Reasoning\\ for Natural Language Understanding}
\title{On the Possibilities and Limitations of Multi-hop Reasoning\\ Under Linguistic Imperfections}

\author{
  Daniel Khashabi$^{1}$\thanks{~~ Work was done when the first author was affiliated with the University of Pennsylvania.}\ \ \;
  Erfan Sadeqi Azer$^{2}$\ \ \;
  Tushar Khot$^{1}$\ \ \; 
  Ashish Sabharwal$^{1}$\ \ \;
  Dan Roth$^{3}$  \\
  \\
  $^1$Allen Institute for AI, Seattle, USA\\
  $^2$Indiana University, Bloomington, USA\\
  $^3$University of Pennsylvania, Philadelphia, USA
}

\date{}

\begin{document}

\maketitle

\begin{abstract}
Systems for language understanding have become remarkably strong at overcoming linguistic imperfections in tasks involving phrase matching or simple reasoning. Yet, their accuracy drops dramatically as the number of reasoning steps increases. We present the first formal framework to study such empirical observations. It allows one to \emph{quantify the amount and effect} of ambiguity, redundancy, incompleteness, and inaccuracy that the use of language introduces when representing a hidden conceptual space. The idea is to consider two interrelated spaces: a conceptual \emph{meaning space} that is unambiguous and complete but hidden, and a \emph{linguistic space} that captures a noisy grounding of the meaning space in the words of a language---the level at which all systems, whether neural or symbolic, operate. Applying this framework to a special class of multi-hop reasoning, namely the \emph{connectivity problem} in graphs of relationships between concepts, we derive rigorous intuitions and impossibility results even under this simplified setting. For instance, if a query requires a moderately large (logarithmic) number of hops in the meaning graph, no reasoning system operating over a noisy graph grounded in language is likely to correctly answer it. This highlights a fundamental barrier that extends to a broader class of reasoning problems and systems, and suggests an alternative path forward: focusing on aligning the two spaces via richer representations, before investing in reasoning with many hops.
\end{abstract}



\section{Introduction}

\begin{figure}
    \centering
    \includegraphics[trim=0.31cm 0.05cm 0cm 0.6cm, clip=false, scale=0.43]{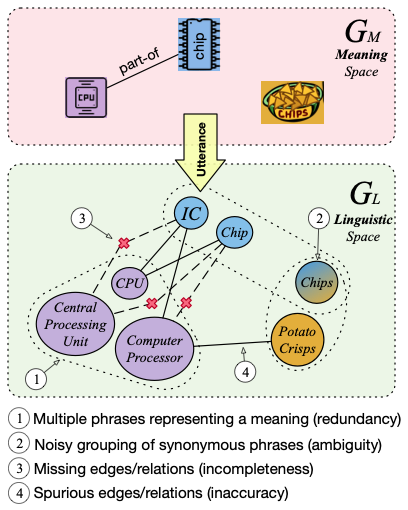}
    \caption{
    The interface between meanings and words capturing linguistic imperfections: each \colorbox{lightred}{meaning} (top) can be \colorbox{lightyellow}{\emph{uttered}} in many ways as \colorbox{lightgreen}{words} (bottom),
    and the same \colorbox{lightgreen}{word} can have
    multiple \colorbox{lightred}{meanings}.
    }
    \label{fig:intro-figure}
\end{figure}


%
Reasoning can be viewed as the process of combining facts and beliefs, in order to infer new conclusions~\cite{johnson1980mental}. While there is a rich literature on reasoning, there is limited understanding of it in the context of natural language, particularly in the presence of linguistic noise.
%
%
There remains a substantial gap between the strong empirical performance and little theoretical understanding of models for common language tasks, such as question answering, reading comprehension, and textual entailment. We focus on \emph{multi-hop or multi-step reasoning over text}, where one must take multiple steps and combine multiple pieces of information to arrive at the correct conclusion. Remarkably, the accuracy of even the best multi-hop models is empirically known to drop drastically as the number of required hops increases~\cite{Yang2018HotpotQAAD,openqa2018}. This is generally viewed as a consequence of linguistic imperfections, such as those illustrated in Figure~\ref{fig:intro-figure}.

We introduce a novel formalism that quantifies these imperfections and makes it possible to theoretically analyze a subclass of multi-hop reasoning algorithms. Our results provide formal support for existing empirical intuitions about the limitations of reasoning with inherently noisy language. Importantly, our formalism can be applied to both neural as well as symbolic systems, as long as they operate on natural language input.

The formalism consists of (A) an abstract model of linguistic knowledge, and (B) a reasoning model. The abstract model is built around the notion of two spaces (see Figure~\ref{fig:intro-figure}): a hidden \meaningspace of unambiguous concepts and an associated \linguisticspace capturing their imperfect grounding in words. The reasoning model captures the ability to infer a property of interest in the \meaningspace (e.g., the relationship between two concepts) while observing only its imperfect representation in the \linguisticspace.

The interaction between the \meaningspace and the \linguisticspace can be seen as simulating and quantifying the \emph{symbol-grounding problem}~\cite{harnad1990symbol}, a key difficulty in operating over language that arises when one tries to accurately map words into their underlying meaning representation. Practitioners often address this challenge by enriching their representations; for example by mapping textual information to Wikipedia entries~\cite{mihalcea2007wikify,RRDA11}, grounding text to executable rules via semantic parsing~\cite{reddy2017universal}, or incorporating contexts into representations~\cite{Peters2018DeepCW,Devlin2018BERTPO}. Our results indicate that advances in higher fidelity representations are key to making progress in multi-hop reasoning.

There are many flavors of multi-hop reasoning, such as finding a particular path of labeled edges through the linguistic space (e.g., textual multi-hop reasoning) or performing discrete operations over a set of such paths (e.g., semantic parsing). In this first study, we explore a common primitive shared by various tasks, namely, the \emph{connectivity problem}: Can we determine whether there is a path of length $k$ between a pair of nodes in the \meaninggraph, while observing only its noisy grounding as the \linguisticgraph?
This simplification clarifies the exposition and analysis; we expect similar impossibility results, as the ones we derive, to hold for a broader class of tasks that rely on connectivity.

We note that even this simplified setting by itself captures interesting cases. Consider the task of inferring the \emph{Is-a} relationship, by reasoning over a noisy taxonomic graph extracted from language. For instance, is it true that
``a chip'' \emph{Is-a} ``a nutrient''?
When these elements cannot be directly looked up in a curated resource such as WordNet~\cite{miller1995wordnet} (e.g., in low-resource languages, for scientific terms, etc.), recognizing the \emph{Is-a} relationship between ``a chip'' and ``a nutrient'' would require finding a path in the linguistic space connecting ``a chip'' to ``a nutrient'', while addressing the associated word-sense issues. \emph{Is-a} edges would need to be extracted from text, resulting in potentially missing and/or extra edges. This corresponds precisely to the linguistic inaccuracies modeled and quantified in our framework.

\paragraph{Contributions.}
This is the first mathematical study of the challenges and limitations of reasoning algorithms in the presence of the symbol-meaning mapping difficulty. Our main contributions are:

\noindent
\textbf{1. Formal Framework.} We establish a novel, linguistically motivated formal framework for analyzing the problem of reasoning about the ground truth (the meaning space) while operating over its noisy linguistic representation. This allows one to derive rigorous intuitions about what various classes of reasoning algorithms \emph{can} and \emph{cannot} achieve.
\noindent
\textbf{2. Impossibility Results.} We study in detail the \emph{connectivity reasoning} problem, focusing on the interplay between linguistic noise (redundancy, ambiguity, incompleteness, and inaccuracy) and the distance (inference steps, or hops) between two concepts in the meaning space. We prove that under low noise, reliable connectivity reasoning is indeed possible up to a few hops (Theorem~\ref{thm:reasoning:possiblity:detailed-informal}). In contrast, even a moderate increase in the noise level makes it provably impossible to assess the connectivity of concepts if they are a logarithmic distance apart in terms of the meaning graph nodes
(Theorems~\ref{thm:multi-hop:impossiblity-informal} and~\ref{thm:impossibility:general:reasoning-informal}). This helps us understand empirical observations of ``semantic drift'' in systems, causing a substantial performance drop beyond 2-3 hops~\cite{fried2015higher-order,jansen2017study}.
We discuss \textbf{practical lessons} from our findings, such as focusing on richer representations and higher-quality abstractions to reduce the number of hops.

\noindent
\textbf{3. Empirical Illustration.} 
While our main contribution is theoretical, we use the experiments to illustrate the implications of our results in a controlled setting. Specifically, we apply the framework to a subset of a real-world knowledge-base, FB15k237~\cite{Toutanova2015ObservedVL}, treated as the meaning graph, illustrating how key parameters of our analytical model influence the possibility (or impossibility) of accurately solving the connectivity problem.

\begin{figure*}
    \centering
    \includegraphics[trim=0cm 0.5cm 0cm 0.2cm, clip=false, width=\textwidth]{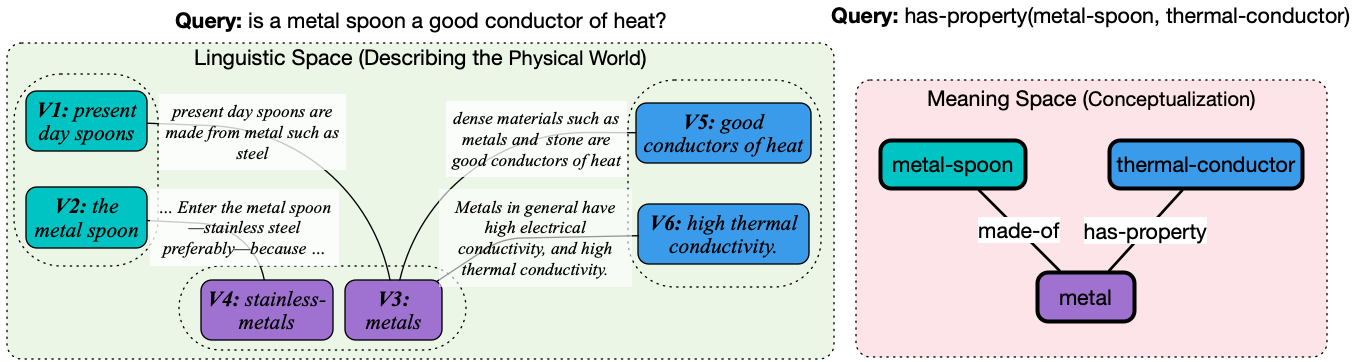}
    \caption{The \meaninggraph (right) captures a clean and unique (canonical) representation of concepts and facts, while the \linguisticgraph (left) contains a noisy, incomplete, and redundant representation of the same knowledge. Shown here are samples of meaning graph nodes and the corresponding linguistic graph nodes for answering the question: \textit{Is a metal spoon a good conductor of heat?}. Relationships in the linguistic graph are here indirectly represented via sentences.}
    \label{fig:ny-example}
\end{figure*}

\section{Overview of the Formalism}


\paragraph{(A) Linguistically-inspired abstract model.}  
We propose an abstract model of linguistic knowledge built around the notion of two spaces (see Figures~\ref{fig:intro-figure} and~\ref{fig:ny-example}). The \meaningspace refers to the internal conceptualization in human mind, where we assume the information is free of noise and uncertainty. In contrast to human thinking that happens in this noise-free space, \emph{human expression} of thought via the \utterance of language introduces many imperfections, and happens in the \linguisticspace. This linguistic space is often redundant (e.g., multiple words\footnote{For simplicity of exposition, we use the term `words' throughout this paper as the unit of information in the linguistic space. Often, the unit of information is instead a short phrase. Our formalism continues to apply to this case as well.} such as ``CPU'' and ``computer processor'' express the same meaning), ambiguous (e.g., a word like ``chips'' could refer to multiple meanings ), incomplete (e.g., common-sense relations never expressed in text), and inaccurate (e.g., incorrect facts written down in text). Importantly, the noisy linguistic space---with its redundancy, ambiguity, incompleteness, and inaccuracy---is what a machine reasoning algorithm operates in.


\paragraph{(B) Reasoning model.}
For the purposes of this work, we define reasoning as the ability to infer the existence or absence of a property of interest in the \meaningspace, by only observing its imperfect representation in the \linguisticspace. Specifically, we focus on properties that can be captured by graphs, namely the \meaninggraph which connects concepts via semantic relationships, and the (noisy) \linguisticgraph which connects words via language. Nodes in the linguistic graph may represent words in various ways, such as using symbols~\cite{miller1995wordnet}, fixed vectors~\cite{mikolov2013}, or even words in context as captured by contextual vectors~\cite{Peters2018DeepCW,Devlin2018BERTPO}.

The target property in the meaning graph characterizes the reasoning task. E.g., 
\emph{relation extraction} corresponds to determining the relationship between two meaning graph nodes, by observing their grounding in text as linguistic graph nodes.
\emph{Word sense disambiguation} (WSD) corresponds to identifying a meaning graph node given a word (a node in the linguistic graph) and its surrounding context. 
\emph{Question-answering} may ask for a linguistic node that has a desired semantic relationship (when mapped to the meaning space) to another given node (e.g., what is the capital of X?).

\subsection{Example}
Figure~\ref{fig:ny-example} illustrates a reasoning setting that includes edge semantics. Most humans understand that \emph{\textbf{V1}:``present day spoons''} and \emph{\textbf{V2}:``the metal spoons''} are equivalent nodes (have the same meaning) for the purposes of the query, ``is a metal spoon a good conductor of heat?''. However, a machine must infer this.
In the linguistic space (left), the semantics of the connections between nodes are expressed through natural language sentences, which also provide a context for the nodes. 
For example, edges in the \meaninggraph (right) directly express the semantic relation {\tt \small \textbf{has-property(metal, thermal-conductor)}}, while a machine, operating on language, may struggle to infer this from reading Internet text expressed in various ways, e.g.,
\emph{``dense materials such as [\textbf{V3:}]metals and stones are [\textbf{V5:}]good conductors of heat''}.

To ground this in existing efforts, consider \emph{multi-hop} reasoning for QA systems~\cite{tableilp2016:ijcai,Jansen2018WorldTreeAC}. Here the reasoning task is to connect local information, via multiple local ``hops'', in order to arrive at a conclusion. In the meaning graph, one can trace a path of locally connected nodes to verify the correctness of a query; for example, the query {\tt \small \textbf{has-property(metal-spoon, thermal-conductor)}}
can be verified by tracing a sequence of nodes in Fig.~\ref{fig:ny-example}. 
Thus, answering some queries can be cast as inferring the existence of a path connecting two nodes.\footnote{This particular grounding is meant to help relate our graph-based formalism to existing applications, and is 
not the only way of realizing reasoning on graphs.}
While doing so on the \meaninggraph is straightforward, reliably doing so on the noisy \linguisticgraph is not. Intuitively, each local ``hop'' introduces additional noise, allowing reliable inference to be performed only when it does not require too many steps in the underlying \meaninggraph. To study this issue, our approach quantifies the effect of noise accumulation for long-range reasoning.

\section{Related Work}

%
AI literature contains a variety of formalisms for automated reasoning,
including reasoning with logical representations~\cite{mccarthy1963programs}, semantic networks~\cite{quillan1966semantic}, 
Bayesian networks~\cite{Pearl1988ProbabilisticReasoning}, etc.
For our analysis, we use a graphical formalism since it provides a useful modeling tool and is a commonly used abstraction for representing items and relations between them. 
One major challenge in reasoning with natural language tasks is grounding  free-form text to a higher-level meaning. 
Example proposals to deal with this challenge include semantic parsing~\cite{steedman2011combinatory}, 
linking to a knowledge base~\cite{mihalcea2007wikify},
and mapping to semantic frames~\cite{PunyakanokRoYi04a}. These methods can be viewed as approximate solutions
to the \emph{symbol-grounding problem}~\cite{harnad1990symbol,taddeo2005solving} --- the challenge of mapping symbols to a representation that reflect their accurate meaning.  
Our formalism is an effort to better understand the fundamental difficulties that this problem poses. 

Several efforts focus on reasoning with disambiguated inputs, such as using executable formulas~\cite{reddy2017universal,Angeli2014NaturalLINL} and chaining relations to infer new relations~\cite{Socher2013ReasoningWN}. 
Our analysis covers any algorithm for inferring patterns that can be represented in a graphical form, e.g.,  
chaining local information, often referred to as \emph{multi-hop} reasoning~\cite{jansen2016s,LinRX2018MultiHopKG}. 
For example, \namecite{jansen2017framing} propose a structured multi-hop reasoning approach by aggregating sentential information from multiple KBs. They demonstrate improvements with few hops and degradation when aggregating more than 2-3 sentences.
Our theoretical findings align well with such observations and further quantify them.




\newcommand{\gm}{\text{{$G_M$}}\xspace}
\newcommand{\gl}{\text{{$G_L$}}\xspace}

\newcommand{\vm}{\text{$V_M$}}
\newcommand{\vs}{\text{$V_L$}}

\newcommand{\emm}{\text{$E_M$}}
\newcommand{\es}{\text{$E_L$}}

\newcommand{\Gm}{\text{{$G_M$}}(\vm, \emm)\xspace}
\newcommand{\Gl}{\text{{$G_L$}}(\vs, \es)\xspace}

\newcommand{\oracle}{\text{$\mathcal{O}$}}
\newcommand{\noisy}{\text{$\mathcal{F}$}}
\newcommand{\radius}{\text{$\mathcal{R}$}}

\newcommand{\perf}{\text{$\mathcal{P}$}}
\newcommand{\bernoulli}{\text{Bern}}

\section{The Meaning-Language Interface}
\label{sec:framework}

We formalize our conceptual and linguistic notions of knowledge spaces as undirected graphs:
\begin{itemize}[leftmargin=.25cm]
\itemsep0em 
    \item The \meaningspace, $M$, is a conceptual hidden space where all the facts are accurate and complete, without ambiguity. 
    We focus on the knowledge in this space that can be represented as an undirected graph, \Gm.
    This knowledge is hidden, and representative of the information that exists within human minds.

    \item The \linguisticspace, $L$, is the space of knowledge represented in natural language for machine consumption (written sentences, curated knowledge-bases, etc.). We assume access to a graph \Gl\ in this space that is a noisy approximation of $G_M$.
\end{itemize}

The two spaces interact: when we read a sentence, we are reading from the \emph{linguistic space} and interpreting it in the \emph{meaning space}. 
When writing out our thoughts, we symbolize our thought process, by moving them from \emph{meaning space} to the \emph{linguistic space}. 
Fig.\ref{fig:intro-figure} provides a high-level view of this interaction.

A reasoning system operates in the linguistic space and is unaware of the exact structure and information encoded in \gm. How well it performs depends on the local connectivity of \gm and the level of noise in \gl, which is discussed next.


\newcommand{\ins}{\mathcal{P}}

\subsection{Meaning-to-Language mapping}
Let $\oracle: V_M \rightarrow  2^{V_L}$ denote an oracle function that captures the set of nodes in \gl that each (hidden) node in \gm should ideally map to in a noise-free setting. This represents multiple ways in which a single meaning can be equivalently represented in language. When $w \in \oracle(m)$, i.e., $w$ is one of the words the oracle maps a meaning $m$ to, we write $\oracle^{-1}(w) = m$ to denote the reverse mapping. For example, if CHIP represents a node in the meaning graph for the common snack food, then \oracle(CHIP) = \{``potato chips'', ``crisps'', ...\}.

Ideally, if two nodes $m_1$ and $m_2$ in $V_M$ have an edge in $G_M$, we would want an edge between every pair of nodes in $\oracle(m_1) \times \oracle(m_2)$ in $G_L$. The actual graph \gl (discussed next) will be a noisy approximation of this ideal case. 

\newcommand{\alg}[1]{\mathsf{ALG}(#1)}

\subsection{Generative Model of Linguistic Graphs}
Linguistic graphs in our framework are constructed via a generative process. Starting with \gm, we sample $G_L \leftarrow \alg{G_M}$ using a stochastic process described in Alg.\ref{al:sampling}. Informally, the algorithm simulates the process of transforming conceptual information into linguistic utterances (web-pages, conversations, knowledge bases, etc.).
This construction captures a few key properties of linguistic representation of meaning via 3 control parameters: \emph{replication factor} $\lambda$, \emph{edge retention probability} $p_+$, and \emph{spurious edge creation probability} $p_-$.

\begin{algorithm}[t]
\small
\KwIn{Meaning graph \Gm, discrete distribution $r(\lambda)$, edge retention prob. $p_+$, edge creation prob. $p_-$}
\KwOut{Linguistic graph \Gl}
 \ForEach{$v \in \vm$}{
     sample $k \sim r(\lambda)$ \\
     construct a collection of new nodes $U$ s.t.\ $|U| = k$ \\
     $\vs \leftarrow \vs  \cup U$,  \hspace{0.2cm} \oracle$(v) \leftarrow U$  \\ 
    
 }
 
 \ForEach{$(m_1,  m_2) \in (\vm \times \vm), m_1 \neq m_2$}{
    $W_1 \leftarrow \oracle(m_1)$, $W_2 \leftarrow \oracle(m_2)$  \\ 
    \ForEach{$e \in W_1 \times W_2$}{
        \eIf{$(m_1,  m_2) \in \emm$}{
            with probability $p_+$: $\es \leftarrow \es  \cup \{ e \} $   
        }{ 
            with probability $p_-$: $\es \leftarrow \es  \cup \{ e \} $
        }
    }
 }
 \caption{
 Generative, stochastic construction of a linguistic graph \gl\ given a meaning graph \gm. 
 }
 \label{al:sampling}
\end{algorithm}

Each node in \gm is mapped to multiple linguistic nodes (the exact number drawn from a distribution $r(\lambda)$, parameterized by $\lambda$), which models \textbf{redundancy} in \gl. \textbf{Incompleteness} of knowledge is modeled by not having all edges of \gm be retained in \gl (controlled by $p_+$). Further, \gl contains spurious edges that do not correspond to any edge in \gm, representing \textbf{inaccuracy} (controlled by $p_-$). The extreme case of noise-free construction of \gl corresponds to having $r(\lambda)$ be concentrated at $1$, $p_+ = 1$, and $p_- = 0$.



\subsection{Noisy Similarity Metric}
Additionally, we include linguistic similarity based edges to model \textbf{ambiguity}, i.e., a single word mapping to multiple meanings. Specifically, we view ambiguity as treating (or confusing) two distinct linguistic nodes for the same word as identical even when they correspond to different meaning nodes (e.g., confusing a ``bat" node for animals with a ``bat'' node for sports).

Similarity metrics are typically used to judge the equivalence of words, with or without context.
Let $\rho: V_L \times V_L \rightarrow \{0, 1\}$ be such a metric, where $\rho(w, w')=1$ for $w,w' \in V_L$ denotes the equivalence of these two nodes in \gl. We define $\rho$ to be a noisy version of the true node similarity as determined by the oracle $\oracle$:
\vspace{-0.05cm}
\begin{align*}
\rho(w, w') \triangleq &
  \begin{cases}
  1 - \text{Bern}(\varepsilon_+) &  \text{if\ } \oracle^{-1}(w) = \oracle^{-1}(w') \\ 
  \text{Bern}(\varepsilon_-) &  \text{otherwise}
  \end{cases}, 
  \vspace{-0.1cm}
\end{align*}
where $\varepsilon_+, \varepsilon_- \in (0, 1)$ are the noise parameters of $\rho$, both typically close to zero, and Bern$(p)$ denotes the Bernoulli distribution with parameter $p$. 

Intuitively, $\rho$ is a \emph{perturbed} version of true similarity (as defined by $\oracle$), with small random noise (parameterized with $\varepsilon_+$ and $\varepsilon_-$). With a high probability $1-\varepsilon_{+/-}$, $\rho$ returns the correct similarity decision as determined by the oracle (i.e., whether two words have the same meaning).
The extreme case of $\varepsilon_+ = \varepsilon_- = 0$ models the perfect similarity metric. In practice, even the best similarity systems are noisy, captured here with $\varepsilon_{+/-} > 0$.

We assume reasoning algorithms have access to $\gl$ and $\rho$, and that they use the following procedure to verify the existence of a direct connection between two nodes: 
\FrameSep0pt
\begin{framed}
  \begin{algorithmic}
        \Function{NodePairConnectivity}{$w, w'$}
         \State \textbf{return} $(w, w') \in E_L$ \ \ or \ \ $\rho(w, w') = 1$ 
        \EndFunction
    \end{algorithmic}
\end{framed}


\newcommand{\tuple}{\mathbf{v_L}}
\newcommand{\tuplem}{\mathbf{v_M}}

Several corner cases result in uninteresting meaning or linguistic graphs. We focus on a regime of ``non-trivial'' cases where \gm is neither overly-connected nor overly-sparse, there is non-zero noise ($p_-, \varepsilon_-, \varepsilon_+ > 0$) and incomplete information ($p_+ < 1$), and noisy content does not dominate actual information (e.g., $p_- \ll p_+$).\footnote{Defn.~\ref{defn:nontrivial} in the Appendix gives a precise characterization.}
Henceforth, we consider only such ``non-trivial'' instances.

Throughout, we will also assume that the replication factor (i.e., the number of linguistic nodes corresponding to each meaning node) is a constant, i.e., $r$ is  such that $\prob{|U| = \lambda} = 1$.

\section{Hypothesis Testing Setup: Reasoning About Meaning via Words}
While a reasoning engine only sees the linguistic graph $G_L$, it must make inferences about the latent meaning graph $G_M$. Given a pair of nodes $\tuple:= \{w, w'\} \subset V_L$, the reasoning algorithm must then predict properties about the corresponding nodes $\tuplem =\{m, m'\} = \{ \oracle^{-1}(w), \oracle^{-1}(w') \}$. 

We use a hypothesis testing setup to assess the likelihood of two disjoint hypotheses defined over \gm, denoted $H^{\circled{1}}_M(\tuplem)$ and $H^{\circled{2}}_M(\tuplem)$ (e.g., whether $\tuplem$ are connected in $\gm$ or not). Given observations $X_L(\tuple)$ about linguistic nodes (e.g., whether $\tuple$ are connected in $\gl$), we define the \emph{goal of a reasoning algorithm} as inferring which of the two hypotheses about \gm\ is more likely to have resulted in these observations, under the sampling process (Alg.\ref{al:sampling}). That is, we are interested in:
\vspace{-0.05cm}
\begin{equation}
\label{eq:decision}
\underset{h \in \{ H^{\circled{1}}_M(\tuplem), 
H^{\circled{2}}_M(\tuplem)\}}{\text{argmax}}  \probTwo{h}{X_L(\tuple)}
\vspace{-0.03cm}
\end{equation}
where $\probTwo{h}{x}$ denotes the probability of an event $x$ in the sample space induced by Alg.\ref{al:sampling}, when (hidden) $\gm$ satisfies hypothesis $h$. Defn.~\ref{defn:reasoning-problem} in the Appendix formalizes this.

Since we start with two disjoint hypotheses on $\gm$, the resulting probability spaces are generally different, making it plausible to identify the correct hypothesis with high confidence. However, with sufficient noise in the sampling process, it can be difficult for an algorithm based on the linguistic graph to distinguish the two resulting probability spaces (corresponding to the two hypotheses), depending on observations $X_L(\tuple)$ used by the algorithm and the parameters of the sampling process. For example, the \emph{distance between linguistic nodes} can often be an insufficient indicator for distinguishing these two hypotheses. We will explore these two contrasting behaviors in the next section.

Not all observations are equally effective in distinguishing $h_1$ from $h_2$. We say $X_L(\tuple)$ \emph{$\gamma$-separates} them if:
\begin{equation}
        \probTwo{ h_1 }{ X_L(\tuple) } - \probTwo{ h_2 }{ X_L(\tuple) } \, \geq \, \gamma.
\label{eqn:gamma-separation}
\end{equation}
(formal definition in Appendix, Defn.~\ref{defn:gamma-separation}) We can view $\gamma$ as the \emph{gap} between the likelihoods of $X_L(\tuple)$ having originated from a meaning graph satisfying $h_1$ vs.\ one satisfying $h_2$. When $\gamma = 1$, $X_L(\tuple)$ is a perfect discriminator for $h_1$ and $h_2$. In general, any positive $\gamma$ bounded away from $1$ yields a valuable observation,\footnote{If the above probability gap is negative, one can instead use the complement of $X_L(\tuple)$ for $\gamma$-separation.} and a reasoning algorithm:
\FrameSep0pt
\begin{framed}
  \small 
  \begin{algorithmic}
        \Function{Separator$\_{X_L}$}{$\gl, \tuple=\{w, w'\}$}
         \State \textbf{if} $X_L(\tuple) = 1$
            {\textbf{then} return $h_1$}
            {\textbf{else} return $h_2$}
        \EndFunction
    \end{algorithmic}
\end{framed}


Importantly, this algorithm does \emph{not} compute the probabilities in Eqs.~(\ref{eq:decision}) and ~(\ref{eqn:gamma-separation}). Rather, it works with a particular instantiation \gl. We refer to such an algorithm $\mathcal{A}$ as \textbf{$\gamma$-accurate} for $h_1$ and $h_2$ if, under the sampling choices of Alg.\ref{al:sampling}, it outputs the `correct' hypothesis with probability at least $\gamma$; that is, for both $i \in \{1,2\}$: 
$\probTwo{h_i}{\mathcal{A} \text{\ outputs\ } h_i} \geq \gamma$. This happens when $X_L$ $\gamma$-separates $h_1$ and $h_2$ (cf.~Appendix, Prop.~\ref{prop:gamma-accurate}).
The rest of the work will explore when one can obtain a $\gamma$-accurate algorithm, using $\gamma$-separation of the underlying observation as an analysis tool.

\section{Main Results (Informal)}

Having formalized a model of the meaning-language interface above, we now present our main findings. The results in this section are intentionally stated in a somewhat informal manner for ease of exposition and to help convey the main messages clearly.
Formal mathematical statements and supporting proofs
are deferred to the Appendix.\footnote{
While provided for completeness and to facilitate follow-up work, details in the Appendix are not crucial for understanding the key concepts and results of this work.}

One simple but often effective approach for reasoning is to focus on connectivity (as described in Fig.~\ref{fig:ny-example}). Specifically, we consider reasoning chains as valid if they correspond to a short path in the meaning space, and invalid if they correspond to disconnected nodes.
Mathematically, this translates into the \textbf{$d$-connectivity reasoning problem}, defined as follows: Given access to a linguistic graph \gl and two nodes $w,w'$ in it, let $m = \oracle^{-1}(w)$ and $m' = \oracle^{-1}(w')$ denote the corresponding nodes in the (hidden) meaning graph \gm. While observing only \gl, can we distinguish between two hypotheses about \gm, namely, $m,m'$ have a path of length $d$ in \gm, vs.\ $m,m'$ are disconnected? Distinguishing between these two hypotheses, namely $h_1 = m \eone m'$ and $h_2 = m \etwo m'$, is the problem we are interested in solving.

The answer clearly depends on how faithfully \gl reflects 
\gm. In the (unrealistic) noise-free case of \gl = \gm, this problem is trivially solvable by computing the shortest path between $w$ and $w'$ in \gl. More realistically, the higher the noise level is in Alg.\ref{al:sampling} and \textsc{NodePairSimilarity}, the harder it will be for any algorithm operating on \gl to confidently infer a property of \gm.

In other words, the redundancy, ambiguity, incompleteness and inaccuracy of language discussed earlier directly impact the capability and limitations of algorithms that perform connectivity reasoning. Our goal is to quantify this intuition using the parameters $p_+, p_-, \lambda, \varepsilon_+, \varepsilon_-$ of our generative model, and derive possibility and impossibility results.

\paragraph{A simple connectivity testing algorithm.}
The first set of results assume the following algorithm: given $w,w'$ with a desired distance $d$ in \gm, it checks whether $w,w'$ have a path of length at most $\tilde{d}$, which is a function of $d$ and the replication factor $\lambda$. If yes, it declares the corresponding meaning nodes $m,m'$ in \gm have a path of length $d$; otherwise it declares them disconnected.\footnote{The simple connectivity algorithm is described formally using the notion of a \textsc{Separator} in Appendix~\ref{sec:possibility-result}.}

\newcommand{\ball}{\text{$\mathcal{B}$}}

Let $a \oplus b$ denote $a + b - ab$, and $\ball(d)$ denote the maximum number of nodes within distance $d$ of any node in \gm. The following result shows that even the simple algorithm above guarantees accurate reasoning (specifically, it correctly infers connectivity in \gm while only observing \gl) for limited noise and small $d$ (i.e., few hops):


\begin{theorem}
[Possibility Result; Informal]
\label{thm:reasoning:possiblity:detailed-informal}
If $p_-, \varepsilon_-, d,$ and $\gamma$ are small enough such that
$(p_- \oplus \varepsilon_-) \cdot \ball^2(d) < \frac{1}{2e\lambda^2n}$,
then there exists a $\gamma \in [0,1]$ that increases with $p_+, \lambda$ and decreases with $\varepsilon_+$,\footnote{See Defn.~\ref{def:accuracy-threshold} in formal results for exact expression for $\gamma$.}
such that the simple connectivity testing algorithm with $\tilde{d} = \lambda \cdot d + \lambda - 1$ correctly solves the $d$-connectivity problem with probability at least $\gamma$.
\end{theorem}

Here, the probability is over the sampling choices of Alg.~\ref{al:sampling} when constructing \gl, and the function \textsc{NodePairConnectivity} for determining node similarity in \gl.

On the other hand, when there is more noise and $d$ becomes even moderately large---specifically, logarithmic in the number of nodes in \gm---then this simple algorithm no longer works even for small values of desired accuracy $\gamma$:


\begin{theorem}
[Impossibility Result \#1; Informal]
\label{thm:multi-hop:impossiblity-informal}
If $p_-$ and $\varepsilon_-$ are large enough such that
$p_- \oplus \varepsilon_- \geq \frac{c}{\lambda n}$
and $d \in \Omega(\log n)$ where $n$ is the number of nodes in \gm, then the simple connectivity algorithm with $\tilde{d} = \lambda d$ \emph{cannot} solve the $d$-connectivity problem correctly with probability $\gamma$ for any $\gamma > 0$.
\end{theorem}

This result exposes an inherent limitation to multi-hop reasoning: even for small values of noise, the diameter of \gl can quickly become very small, namely, logarithmic in $n$ (similar to the \emph{small-world phenomenon} observed by~\namecite{milgram1967six} in other contexts), at which point the above impossibility result kicks in. Our result affirms that if NLP reasoning algorithms are not designed carefully, such macro behaviors will necessarily become bottlenecks, even for relatively simple tasks such as detecting connectivity.

The above result is for the simple connectivity algorithm. One can imagine other ways to determine connectivity in \gm, such as by analyzing the degree distribution of \gl, looking at its clustering structure, etc. Our third finding extends the impossibility result to this general setting, showing that if the noise level is increased a little more (by a logarithmic factor), then \emph{no} algorithm can infer connectivity in \gm by observing only \gl:


\begin{theorem}
[Impossibility Result \#2; Informal]
\label{thm:impossibility:general:reasoning-informal}
If $p_-$ and $\varepsilon_-$ are large enough such that
$p_- \oplus \varepsilon_- > \frac{c \log n}{\lambda n}$
and $d > \log n$ where $n$ is the number of nodes in \gm, then \emph{any} algorithm \emph{cannot} correctly solve the $d$-connectivity problem correctly with probability $\gamma$ for any $\gamma > 0$.
\end{theorem}


This reveals a fundamental limitation, that we may only be able to infer interesting properties of \gm within small, logarithmic sized neighborhoods. We leave the formal counterparts of these results (Theorems~\ref{thm:reasoning:possiblity:detailed}, \ref{thm:multi-hop:impossiblity}, and \ref{thm:impossiblity:general:reasoning}, resp.) to Appendix~\ref{sec:results-formal}, and focus next on a small scale empirical validation to complement our analytical findings.

\section{Empirical Validation}

Our formal analysis thus far offers worst-case bounds for two regions in a fairly large spectrum of noisy sampling parameters for the linguistic space, namely, when $p_- \oplus \varepsilon_-$ and $d$ are either both small (Theorem~\ref{thm:reasoning:possiblity:detailed-informal}), or both large (Theorems~\ref{thm:multi-hop:impossiblity-informal} and \ref{thm:impossibility:general:reasoning-informal}).

This section complements our theoretical findings in two ways: (a) by grounding the formalism empirically into a real-world knowledge graph, (b) by quantifying the impact of sampling parameters on the connectivity algorithm. 
    We use $\varepsilon_- = 0$ for the experiments, but the effect turns out to be identical to using $\varepsilon_- > 0$ for appropriate adjustments of $p_-$ and $p_+$  
    (see Remark~\ref{remark:folding} in the Appendix).	

We consider FB15k237~\cite{Toutanova2015ObservedVL}, a set of $\langle$head, relation, target$\rangle$ triples from FreeBase~\cite{bollacker2008freebase}. For scalability, we use the movies domain subset (relations {\small \tt /film/*}), with 2855 entity nodes and 4682 relation edges. 
We treat this as the meaning graph \gm\ and sample a linguistic graph \gl (via Alg.\ref{al:sampling}) to simulate the observed graph derived from text.

\begin{figure}
    \centering
    \includegraphics[scale=0.17,trim=1.9cm 1.5cm 0.5cm 2.5cm, clip=false]{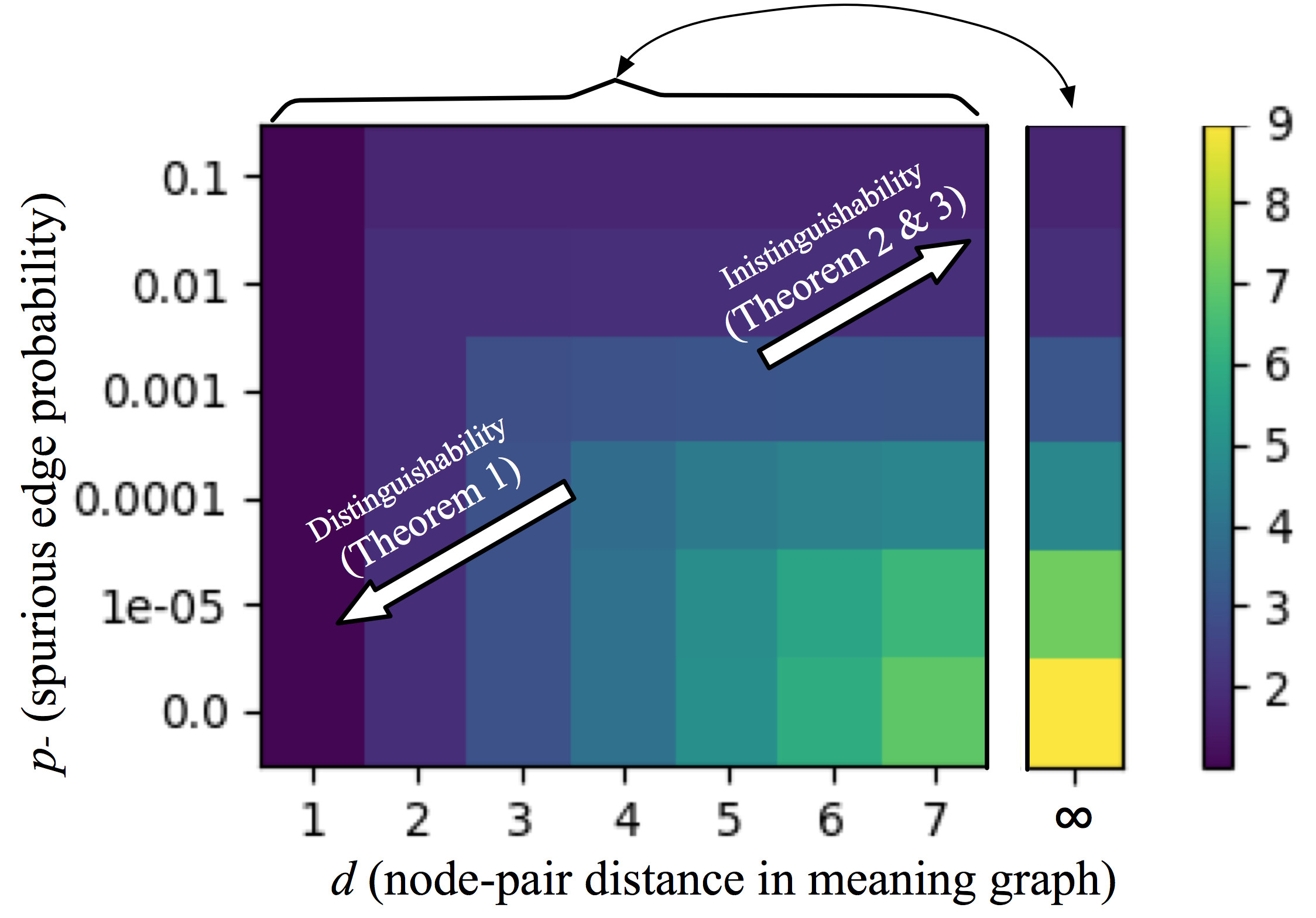}
    \caption{ 
    \small
    Colors in the figure depict the average distance between node-pairs in \gl, for each true distance $d$ (x-axis) in \gm, as the noise parameter $p_-$ (y-axis) is varied. The goal is to distinguish squares in the column for a particular $d$ with the corresponding squares in the right-most column, which corresponds to node-pairs being disconnected. This is easy in the bottom-left regime and becomes progressively harder as we move upward (more noise) or rightward (higher distance in \gm). ($\varepsilon_+=0.7, \lambda = 3$)}
    \label{fig:emp:distance:pm}
\end{figure}

We sample \gl\ for various values of $p_-$ and plot the resulting distances in \gl\ and \gm\ in Fig.~\ref{fig:emp:distance:pm}, as follows. For every value of $p_-$ ($y$-axis), we sample pairs of points in \gm\ that are separated by distance $d$ ($x$-axis). 
For these pairs of points, we compute the average distance between the corresponding linguistic nodes (in sampled \gl), and plot that in the heat map using color shades.

We make two observations from this simulation. First, for lower values of $p_-$, disconnected nodes in \gm\ (rightmost column) are clearly distinguishable from meaning nodes with short paths (small $d$) as predicted by Theorem~\ref{thm:reasoning:possiblity:detailed}, but harder to distinguish from nodes at large distances (large $d$). Second, and in contrast, for higher values of $p_-$, almost every pair of linguistic nodes is connected with a very short path (dark color), making it impossible for a distance-based reasoning algorithm to confidently assess $d$-connectivity in \gm. This simulation also confirms our finding in Theorem~\ref{thm:multi-hop:impossiblity}: any graph with $p_- \geq 1/\lambda n$ (which is $\sim 0.0001$ here) cannot distinguish disconnected meaning nodes from nodes with paths of short (logarithmic) length (top rows).

\section{Implications and Practical Lessons}

Our analysis is motivated by empirical observations of ``semantic drift'' of reasoning algorithms, as the number of hops is increased. For example, \namecite{fried2015higher-order} show modest benefits up to 2-3 hops, and then decreasing performance; \namecite{Jansen2018WorldTreeAC} 
make similar observations in graphs built out of larger structures such as sentences, where the performance drops off around 2 hops. This pattern has interestingly been observed under a variety of representations, including word-level input, graphs, and traversal methods. 
A natural question is whether 
the field might be hitting a fundamental limit on multi-hop information aggregation. 
Our ``impossibility'' results are reaffirmations of the empirical intuition in the field. This means that multi-hop inference (and any algorithm that can be cast in that form), as we've been approaching it, is exceptionally unlikely to breach the few-hop barrier predicted in our analysis. 


There are at least two \textbf{practical lessons}: 
    First, our results suggest that ongoing efforts on ``very long'' multi-hop reasoning, especially without a careful understanding of the limitations, are unlikely to succeed, unless some fundamental building blocks are altered.
    Second, this observation suggests that practitioners must focus on richer representations that allow reasoning with only a ``few'' hops. This, in part, requires higher-fidelity abstraction and grounding mechanisms. It also points to alternatives, such as offline KB completion/expansion mechanisms, which indirectly reduce the number of steps needed at inference time.
    
    Our formal results quantify how different linguistic imperfection parameters affect multi-hop reasoning differently. This provides an informed design choice for which noise parameter to try to control when creating a new representation.

\section{Conclusion}

This work is the first attempt to develop a formal framework for understanding the behavior of complex natural language reasoning in the presence of linguistic noise. 
The importance of this work is two-fold. First, it proposes a novel graph-theoretic paradigm for studying reasoning, inspired by the symbol-meaning problem in the presence of redundancy, ambiguity, incompleteness, and inaccuracy of language. 
Second, it shows how to use this framework to analyze a class of reasoning algorithms. 
%
We expect our findings, as well as those from future extensions to other classes of reasoning algorithms, to have important implications on how we study problems in language comprehension.

\bibliography{ref,ref-latest}
\bibliographystyle{acl_natbib}

\newpage
\input{appendix}

\end{document}

%% file: appendix.tex
\appendix

\onecolumn

\begin{center}
{\LARGE \textbf{Supplemental Material}}
\end{center}

To focus the main text on key findings and intuitions, we abstract away many technical details. Here instead we go into the formal proofs with the necessary mathematical rigor. 
We start with our requisite definitions (\S\ref{appendix:sec:definitions}), followed by formal statements of our three results (\S\ref{sec:results-formal}), and their proofs (\S\ref{sec:supp:proofs}, \S\ref{appendix:proof:connectivity:reasoning}, \S\ref{appendix:proof:general:reasoning}). We conclude the supplementary material with additional empirical results (\S\ref{appendix:empirical:details}).

\newcommand{\pois}[1]{\text{Pois}(#1)}
\newcommand{\normal}[1]{\mathcal{N}(#1)}
\newcommand{\bin}[1]{\text{Bin}(#1)}
\newcommand{\bern}[1]{\text{Bern}(#1)}

\section{Definitions: Reasoning Problems, Hypothesis Testing Setup, and $\gamma$-Separation}
\label{appendix:sec:definitions}

\paragraph{Notation.}
Throughout, we follow the standard notation for asymptotic comparison of functions: 
$O(.)$, $o(.)$, $\Theta(.)$, $\Omega(.),$ and $\omega(.)$~\cite{cormen2009introduction}.
$X \sim f(\theta)$ denotes a random variable $X$ distributed according to probability distribution $f(\theta)$, paramterized by $\theta$.
$\bern{p}$ and $\bin{n,p}$ denote the Bernoulli and Binomial distributions, respectively.
Given random variables $X\sim\bern{p}$ and $Y\sim\bern{q}$, their disjunction $X \vee Y$ is distributed as $\bern{p \oplus q }$, where  
$p \oplus q \triangleq 1 - (1-p)(1-q) = p + q - pq$. We  make extensive use of this notation. 

Let $\dist(u, v)$ be the distance between nodes $u$ and $v$ in $G$. 
A \emph{simple path} (henceforth referred to as just a \emph{path}) is a sequence of adjacent nodes that does not have repeating nodes. Let $u \pathk{d} v$ denote the existence of a path of length $d$ between $u$ and $v$. Similarly, $u \npathk{} v$ denotes $u$ and $v$ are disconnected. The notion of $d$-\emph{neighborhood} is useful when analyzing local properties of graphs:  
\begin{definition}
For a graph $G = (V,E)$, $w \in V$, and $d \in \mathbb{N}$, the \emph{$d$-neighborhood} of $w$ is $\{v \mid \dist(w,v) \leq d\}$, i.e., the `ball' of radius $d$ around $w$. $\ball(w,d)$ denotes the number of nodes in this $d$-neighborhood, and $\ball(d) = \max_{w \in V} \ball(w,d)$. 
\end{definition}

\newcommand{\renymodel}{\mathsf{G(n, p)}}


For the rest of this section we formally the key concepts used in the main text, as well as those necessary for our formal proofs. 
\begin{definition}[Nontrivial Graph Instances]
\label{defn:nontrivial}
    A pair ($\gm, \gl$) of a meaning graph and a linguistic graph created from it is \emph{non-trivial} if its generation process satisfies the following: 
    \begin{enumerate}
        \itemsep0em 
        \item non-zero noise, i.e., $p_-, \varepsilon_-, \varepsilon_+ > 0$; 
        
        \item incomplete information, i.e., $p_+ < 1$;
        
        \item noise content does not dominate the actual information, i.e., $p_- \ll p_+$, $\varepsilon_+ < 0.5$, and $p_+ > 0.5$; 
        
        \item \gm\ is not overly-connected, i.e., $\ball(d) \in o(n)$, where $n$ is the number of nodes in $\gm$;
        
        \item \gm\ is not overly-sparse, i.e., $|E_{G_M}| \in \omega(1)$. 
    \end{enumerate}
\end{definition}


While the reasoning engine only sees the linguistic graph $G_L$, it must make inferences about the potential latent meaning graph. Given a pair of nodes $\tuple:= \{w, w'\} \subset V_L$ in the linguistic graph, the reasoning algorithm must then predict properties about the corresponding nodes $\tuplem =\{m, m'\} = \{ \oracle^{-1}(w), \oracle^{-1}(w') \}$ in the meaning graph. 

We use a hypothesis testing setup to assess the likelihood of two disjoint hypotheses defined over \gm:
$H^{\circled{1}}_M(\tuplem)$ and $H^{\circled{2}}_M(\tuplem)$. Given observations $X_L(\tuple)$ about linguistic nodes, the \emph{goal of a reasoning algorithm} here is to identify which of the two hypotheses about \gm\ is more likely to have resulted in these observations, under the sampling process of Algorithm~\ref{al:sampling}. That is, we are interested in:
\vspace{-0.05cm}
\begin{equation}
\label{eq:decision-appendix-copy}
\underset{h \in \{ H^{\circled{1}}_M(\tuplem), 
H^{\circled{2}}_M(\tuplem)\}}{\text{argmax}}  \probTwo{h}{X_L(\tuple)}
\end{equation}
\vspace{-0.03cm}
where $\probTwo{h}{x}$ denotes the probability of an event $x$ in the sample space induced by Algorithm~\ref{al:sampling}, when the (hidden) meaning graph $\gm$ satisfies hypothesis $h$. Formally:

\begin{definition}[Reasoning Problem]
\label{defn:reasoning-problem}
    The input for an instance $\ins$ of the \emph{reasoning problem} is a collection of parameters that characterize how a linguistic graph $\gl$ is generated from a (latent) meaning graph $\gm$, two hypotheses $H^{\circled{1}}_M(\tuplem), H^{\circled{2}}_M(\tuplem)$ about $\gm$, and available observations $X_L(\tuple)$ in $\gl$.
    The reasoning problem, $\ins(p_+$, $p_-$, $\varepsilon_+$, $\varepsilon_-$, $\ball(d)$, n, $\lambda$, $H^{\circled{1}}_M(\tuplem)$, $H^{\circled{2}}_M(\tuplem)$, $X_L(\tuple))$, is to map the input to the hypothesis $h$ as per Eq.~(\ref{eq:decision}).
\end{definition}

Since we start with two disjoint hypotheses on $\gm$, the resulting probability spaces are generally different, making it plausible to identify the correct hypothesis with high confidence. On the other hand, with sufficient noise in the sampling process, it can also become difficult for an algorithm to distinguish the two resulting probability spaces (corresponding to the two hypotheses),  especially depending on the observations $X_L(\tuple)$ used by the algorithm and the parameters of the sampling process. For example, the \emph{distance between linguistic nodes} can often be an insufficient indicator for distinguishing these two hypotheses. We will explore these two contrasting reasoning behaviors in the next section.

We use ``separation'' to measure how effective is an observation $X_L$ in distinguishing between the two hypotheses:
\vspace{-0.05cm}
\begin{definition}[$\gamma$-Separation]
\label{defn:gamma-separation}
    For $\gamma \in [0,1]$ and a reasoning problem instance $\ins$ with two hypotheses $h_1 = H^{\circled{1}}_M(\tuplem)$ and $h_2 = H^{\circled{2}}_M(\tuplem)$, we say an observation $X_L(\tuple)$ in the linguistic space \emph{$\gamma$-separates} $h_1$ from $h_2$ if: 
    $$
        \probTwo{ h_1 }{ X_L(\tuple) } - \probTwo{ h_2 }{ X_L(\tuple) } \, \geq \, \gamma.
    $$
\end{definition}
\vspace{-0.05cm}
We can view $\gamma$ as the \emph{gap} between the likelihoods of the observation $X_L(\tuple)$ having originated from a meaning graph satisfying hypothesis $h_1$ vs.\ one satisfying hypothesis $h_2$. When $\gamma = 1$, $X_L(\tuple)$ is a perfect discriminator for distinguishing $h_1$ and $h_2$. In general, any positive 
$\gamma$ bounded away from $1$ yields a valuable observation.\footnote{If the above probability gap is negative, one can instead use the complement of $X_L(\tuple)$ for $\gamma$-separation.}

Given an observation $X_L$ that $\gamma$-separates $h_1$ and $h_2$, there is a simple algorithm that distinguishes $h_1$ from $h_2$:
%
\begin{framed}
  \begin{algorithmic}
        \Function{Separator$_{X_L}$}{$\gl, \tuple=\{w, w'\}$}
         \State \textbf{if} $X_L(\tuple) = 1$
            {\textbf{then} return $h_1$}
            {\textbf{else} return $h_2$}
        \EndFunction
    \end{algorithmic}
\end{framed}

Importantly, this algorithm does \emph{not} compute the probabilities in Definition~\ref{defn:gamma-separation}. Rather, it works with a particular instantiation \gl\ of the linguistic graph. We refer to such an algorithm $\mathcal{A}$ as \textbf{$\gamma$-accurate} for $h_1$ and $h_2$ if, under the sampling choices of Algorithm~\ref{al:sampling}, it outputs the `correct' hypothesis with probability at least $\gamma$; that is, for both $i \in \{1,2\}$: 
$\probTwo{h_i}{\mathcal{A} \text{\ outputs\ } h_i} \geq \gamma.$

\begin{proposition}
\label{prop:gamma-accurate}
If observation $X_L$ $\gamma$-separates $h_1$ and $h_2$, then algorithm \textsc{Separator}$_{X_L}$ is $\gamma$-accurate for $h_1$ and $h_2$.
\end{proposition}

\begin{proof}
Let $\mathcal{A}$ denote \textsc{Separator}$_{X_L}$ for brevity. Combining $\gamma$-separation of $X_L$ with how $\mathcal{A}$ operates, we obtain:
{
\small
\begin{align*}
    \probTwo{h_1}{\mathcal{A} \text{\ outputs\ } h_1} - \probTwo{h_2}{\mathcal{A} \text{\ outputs\ } h_1} & \geq \gamma \\
    \Rightarrow \probTwo{h_1}{\mathcal{A} \text{\ outputs\ } h_1} + \probTwo{h_2}{\mathcal{A} \text{\ outputs\ } h_2} & \geq 1 + \gamma
\end{align*}
}
Since each term on the left is bounded above by $1$, each of them must also be at least $\gamma$, as desired.
\end{proof}

The rest of the work will explore when one can obtain a $\gamma$-accurate algorithm, using $\gamma$-separation of the underlying observation as an analysis tool.

\changed{
The following observation allows a simplification of the proofs, without loss of any generality.
}

\begin{remark}
\label{remark:folding}
Since our procedure doesn't treat similarity edges and meaning-to-symbol noise edges differently, we can `fold' $\varepsilon_-$ into $p_-$ and $p_+$ (by increasing edge probabilities). 
More generally, the results are identical whether one uses $p_+, p_-, \varepsilon_-$ or $p'_+, p'_-, \varepsilon'_-$, as long as: 
$$
\begin{cases}
p_+ \oplus \varepsilon_- = p'_+ \oplus  \varepsilon'_- \\ 
p_- \oplus \varepsilon_- = p'_- \oplus \varepsilon'_-
\end{cases}
$$ 
For any $p_+$ and $\varepsilon_-$, we can find a $p'_+$ such that $\varepsilon'_-$ = 0. Thus, w.l.o.g., in the following analysis we derive results only using $p_+$ and $p_-$ (i.e. assume $\varepsilon'_-$ = 0). Note that we  expand these terms to $p_+ \oplus \varepsilon_-$ and $p_- \oplus \varepsilon_-$ respectively in the final results.
\end{remark}

\clearpage

\section{Main Results (Formal)}
\label{sec:results-formal}

This section presents a more formal version of our results. As discussed below, the formal theorems are best stated using the notions of hypothesis testing, observations on linguistic graphs \gl, and $\gamma$-separation of two hypotheses about \gm using these observations on \gl.

\subsection{Connectivity Reasoning Algorithm}
\label{sec:possibility-result}

For nodes $m, m'$ in $\gm$, we refer to distinguishing the following two hypotheses as the \textbf{$d$-connectivity reasoning problem} and find that even these two extreme worlds can be difficult to separate:
\vspace*{-0.2cm}
$$h_1 = m \eone m'\text{,\ \ and\ \ }\; h_2 = m \etwo m'$$


For reasoning algorithms, a natural \emph{test} is the connectivity of linguistic nodes in $\gl$ via the \textsc{NodePairConnectivity} function. Specifically, checking whether there is a path of length $\leq \tilde{d}$ between $w, w'$: 
 \vspace*{-0.2cm}
$$
 X^{\tilde{d}}_L(w, w') = \; w \pathk{\leq \tilde{d}} w'
$$
Existing multi-hop reasoning models~\cite{khot2017tupleinf} use similar approaches to identify valid reasoning chains. 

The corresponding \textbf{connectivity algorithm} is \textsc{Separator}$\_{X^{\tilde{d}}_L}$, which we would like to be $\gamma$-accurate for $h_1$ and $h_2$.
Next, we derive bounds on $\gamma$ for these specific hypotheses and observations $X_L(\tuple)$. While the space of possible hypotheses and observations is large, the above natural and simple choices still allow us to derive valuable intuitions. 

\subsubsection{Possibility of Accurate Connectivity Reasoning}

We begin by defining the following accuracy threshold, $\gamma^*$, as a function of the parameters for sampling a linguistic graph:
\begin{definition}
\label{def:accuracy-threshold}
Given $n, d \in \mathbb{N}$ and linguistic graph sampling parameters $p_+, \varepsilon_+, \lambda$, define $ \gamma^*(n, d, p_+, \varepsilon_+, \varepsilon_-, \lambda) $ as
\begin{equation*}\label{model3_coef}
        \left( 1 - \left( 1- (p_+ \oplus \varepsilon_-) \right)^{\lambda^2} \right)^{d} 
       \cdot \left( 1 - 2 e^3 \varepsilon_+ ^{\lambda/2} \right)^{d+1} - 2en (\lambda\ball(d))^2 p_-.
\end{equation*}
\vspace{-0.05cm}
\end{definition}
This expression, although complex-looking, behaves in a natural way. E.g., the accuracy threshold $\gamma^*$ increases (higher accuracy) as $p_+$ increases (higher edge retention) or $\varepsilon_+$ decreases (fewer dropped connections between replicas). Similarly, as $\lambda$ increases (higher replication), the impact of the noise on edges between node clusters
or $d$ decreases (shorter paths), again increasing the accuracy threshold.


%
\begin{theorem}[formal statement of Theorem~\ref{thm:reasoning:possiblity:detailed-informal}]
\label{thm:reasoning:possiblity:detailed}
Let $p_+, p_-, \lambda$ be the parameters of Alg.\ref{al:sampling} on a meaning graph with $n$ nodes. Let $\varepsilon_+, \varepsilon_-$ be the parameters of \textsc{NodePairConnectivity}. Let $d \in \mathbb{N}$ and {$\tilde{d} = \lambda \cdot d + \lambda - 1$.} If
\vspace{-0.10cm}
$$
    \changed{ (p_- \oplus \varepsilon_-) } \cdot \ball^2(d) < \frac{1}{\changed{2e}\lambda^2n},
$$
and $\gamma = \max \{ 0, \gamma^*(n, d, p_+, \varepsilon_+, \varepsilon_-, \lambda) \}$, then algorithm \textsc{Separator}$\_{X^{\tilde{d}}_L}$ is $\gamma$-accurate for $d$-connectivity problem.
\end{theorem}

\subsection{Limits of Connectivity-Based Algorithms}
\label{sec:impossibility-results}

We show that as the distance $d$ between two meaning nodes increases, it becomes difficult to make any inference about their connectivity by assessing connectivity of the corresponding linguistic-graph nodes. 
Specifically, if $d$ is at least logarithmic in the number of meaning nodes, then, even with small noise, the algorithm will see all node-pairs as being within distance $d$, making informative inference unlikely.

\begin{theorem}[formal statement of Theorem~\ref{thm:multi-hop:impossiblity-informal}]
\label{thm:multi-hop:impossiblity}
Let $c > 1$ be a constant and $p_-, \lambda$ be parameters of the sampling process in Alg.\ref{al:sampling} on a meaning graph $\gm$ with $n$ nodes. Let $\varepsilon_-$ be a parameter of \textsc{NodePairConnectivity}. Let $d \in \mathbb{N}$ and $\tilde{d} = \lambda d$. If
$$
    \vspace{-0.01cm}
    \changed{p_- \oplus \varepsilon_-} \geq \frac{c}{\lambda n} \ \ \ \text{and} \ \ \ d \in \Omega (\log n),
    \vspace{-0.01cm}
$$
then the connectivity algorithm \textsc{Separator}$\_{X^{\tilde{d}}_L}$ almost-surely infers any node-pair in $\gm$ as connected, and is thus not $\gamma$-accurate for any $\gamma > 0$ for the $d$-connectivity problem.
\end{theorem}

Note that preconditions of Theorems~\ref{thm:reasoning:possiblity:detailed} and~\ref{thm:multi-hop:impossiblity} are disjoint, that is, both results don't apply simultaneously. Since $\ball(.) \geq 1$ and $\lambda \geq 1$, Theorem~\ref{thm:reasoning:possiblity:detailed} requires $\changed{p_- \oplus \varepsilon_-} 
\leq \changed{ \frac{1}{\changed{2e}\lambda^2 n} < \frac{1}{\lambda^2 n}}$, whereas Theorem~\ref{thm:multi-hop:impossiblity} applies when $\changed{p_- \oplus \varepsilon_-} \geq \frac{c}{\lambda n} > \frac{1}{\lambda^2 n}$.

\subsection{Limits of General Algorithms}


We now extend the result to an algorithm agnostic setting, where no assumption is made on the choice of the \textsc{Separator} algorithm or $X_L(\tuple)$. For instance, $X_L(\tuple)$ could make use of the entire degree distribution of $\gl$, compute the number of disjoint paths between various linguistic nodes, cluster nodes, etc. The analysis uses spectral properties of graphs to quantify local information. 

\begin{theorem}[formal statement of Theorem~\ref{thm:impossibility:general:reasoning-informal}]
\label{thm:impossiblity:general:reasoning}
Let $c > 0$ be a constant and $p_-, \lambda$ be parameters of the sampling process in Alg.\ref{al:sampling} on a meaning graph $\gm$ with $n$ nodes. Let $\varepsilon_-$ be a parameter of \textsc{NodePairConnectivity}. Let $d \in \mathbb{N}$. If
$$
    \changed{p_- \oplus \varepsilon_-} > \frac{c \log n}{\lambda n} \ \ \ \text{and} \ \ \ d > \log n,
$$
then there exists $n_0 \in \mathbb{N}$ s.t.\ for all $n \geq n_0$, \emph{no} algorithm can distinguish, with a high probability, between two nodes in \gm\ having a $d$-path vs.\ being disconnected, and is thus not $\gamma$-accurate for any $\gamma > 0$ for the $d$-connectivity problem.
\end{theorem}

This reveals a fundamental limitation: under noisy conditions, our ability to infer interesting phenomena in the meaning space is limited to a small, logarithmic neighborhood.

\clearpage

\section{Proofs: Possibility of Accurate Connectivity Reasoning}
\label{sec:supp:proofs}

In this section we provide the proofs of the additional lemmas necessary for proving the intermediate results. 
First we introduce a few useful lemmas, and then move on to the proof of Theorem~\ref{thm:reasoning:possiblity:detailed}. 

We introduce the following lemmas which will be used in connectivity analysis of the clusters of the nodes $\oracle(m)$.
\begin{lemma}[Connectivity of a random graph~\cite{gilbert1959random}]
\label{connectivity:probability}
Let $P_n$ denote the probability of the event that a random undirected graph $\renymodel$ ($p > 0.5$) is connected.
This probability can be lower-bounded as following: 
$$
P_n \geq 1 - \left[ q^{n-1} \left\lbrace (1 + q^{(n-2)/2} )^{n-1} - q^{(n-2)(n-1)/2} \right\rbrace + q^{n/2}\left\lbrace (1 + q^{(n-2)/2})^{n-1} - 1\right\rbrace \right],
$$
where $q = 1-p$. 
\end{lemma}

See \namecite{gilbert1959random} for a proof of this lemma. Since $q \in (0, 1)$, this implies that $P_n  \rightarrow 1$ as $n$ increases. The following lemma provides a simpler version of the above probability: 
\begin{corollary}[Connectivity of a random graph~\cite{gilbert1959random}]
\label{cor:connectivity:likelihood}
The random-graph connectivity probability $P_n$ (Lemma~\ref{connectivity:probability}) can be lower-bounded as following: 
$$
P_n \geq 1 - 2e^3q^{n/2}
$$
\end{corollary}
\begin{proof}
We use the following inequality: 
$$ (1 + \frac{3}{n})^n \leq e^3 $$

Given that $q \leq 0.5, n\geq 1$, one can verify that $q^{(n-2)/2} \leq 3/n$. Combining this with the above inequality gives us,  
$(1 + q^{{n-2}/2})^{n-1} \leq e^3 $. 

With this, we bound the two terms within the two terms of the target inequality: 
$$
\begin{cases}
(1 + q^{(n-2)/2} )^{n-1} - q^{(n-2)(n-1)/2} \leq e^3  \\ 
(1 + q^{(n-2)/2})^{n-1} - 1 \leq e^3 
\end{cases}
$$

$$
\left[ q^{n-1} \left\lbrace (1 + q^{(n-2)/2} )^{n-1} - q^{(n-2)(n-1)/2} \right\rbrace + q^{n/2}\left\lbrace (1 + q^{(n-2)/2})^{n-1} - 1\right\rbrace \right] \leq  e^3 q^{n-1}  + e^3 q^{n/2} \leq  2e^3 q^{n/2}
$$
which concludes the proof. 

\end{proof}

We show a lower-bound on the probability of $w$ and $w'$ being connected given the connectivity of their counterpart nodes in the meaning graph. Notice that the choice of $\tilde{d} = (d+1)(\lambda-1)+d= \lambda \cdot d + \lambda - 1$ is calculated as follows: The type of paths from $w$ to $w'$, we consider for the proof, has at most $\lambda - 1$ edges within the nodes corresponding to one single meaning node, and $d$ edges in between. This lemma will be used in the proof of Theorem~\ref{thm:reasoning:possiblity:detailed}:

\begin{lemma}[Lower bound]
\label{lem:LBthm1}
$\prob{  w \pathk{\tilde{d}} w' | m \pathk{d} m' } \geq \left(1 - 2 e^3 \varepsilon_+ ^{\lambda/2}\right)^{d+1} \cdot  \left(1 - (1-p_+)^{\lambda^2}\right)^{d}$.
\end{lemma}
\begin{proof}
We know that $m$ and $m'$ are connected through some intermediate nodes $m_1, m_2, \cdots, m_\ell$ ($\ell < d$). We show a lower-bound on having a path in the linguistic graph between $w$ and $w'$, through clusters of nodes $\oracle(m_1), \oracle(m_2), \cdots, \oracle(m_\ell)$. We decompose this into two events: 
$$
\begin{cases}
e_1[v] & \text{For a given meaning node } v \text{ its cluster in the linguistic graph, } \oracle(v) \text{ is connected. }\\
e_2[v, u] & \text{For any two connected nodes } (u, v) \text{ in the meaning graph, there is at least an edge} \\
 & \text{connecting their clusters } \oracle(u), \oracle(v) \text{ in the linguistic graph.}
\end{cases}
$$
The desired probability can then be refactored as: 
\begin{align*}
\prob{  w \pathk{\tilde{d}} w' | m \pathk{d} m' }
  & \geq   \prob{  \left(\bigcap_{v\in\{w,m_1,\dots,m_\ell,w'\}} e_1[v]\right)\cap \left(\bigcap_{(v,u)\in\{(w,m_1),\dots,(m_\ell,w')\}} e_2[v,u]\right)  } \\
  & \geq \prob{e_1}^{d+1} \cdot \prob{e_2}^d. 
\end{align*}

We split the two probabilities and identify lower bounds for each. Based on Corollary~\ref{cor:connectivity:likelihood},
$\prob{e_1} \geq 1 - 2e^3\varepsilon_+ ^{\lambda/2}$, and as a result $\prob{e_1}^{d+1} \geq  \left(1 - 2e^3 \varepsilon_+ ^{\lambda/2}\right)^{d+1}$. The probability of connectivity between pair of clusters is 
$\prob{e_2}  = 1 - (1-p_+)^{\lambda^2}$. Thus, similarly,  $\prob{e_2}^{d} \geq  \left(1 - (1-p_+)^{\lambda^2}\right)^{d}$. Combining these two, we obtain:
\begin{equation}
\label{eqn:lower-bound}
    \prob{  w \pathk{\tilde{d}} w' | m \pathk{d} m' } \geq \left(1 - 2 e^3 \varepsilon_+ ^{\lambda/2}\right)^{d+1} \cdot  \left(1 - (1-p_+)^{\lambda^2}\right)^{d}
\end{equation}
\end{proof}

The connectivity analysis of $G_L$ can be challenging since the graph is a non-homogeneous combination of positive and negative edges. 
For the sake of simplifying the probabilistic arguments, given linguistic graph $G_L$, we introduce a non-unique simple graph $\tilde{G}_L$ as follows.
\begin{definition}
Consider a special partitioning of $V_G$ such that the $d$-neighbourhoods of $w$ and $w'$ form two of the partitions and the rest of the nodes are arbitrarily partitioned in a way that the diameter of each component does not exceed $\tilde{d}$.
\begin{itemize}
    \item The set of nodes $V_{\tilde{G}_L}$ of $\tilde{G}_L$ corresponds to the aforementioned partitions.
    
    \item There is an edge $(u,v)\in E_{\tilde{G}_L}$ if and only if at least one node-pair from the partitions of $V_G$ corresponding to $u$ and $v$, respectively, is connected in $E_{G_L}$.
\end{itemize}
\end{definition}

In the following lemma we give an upper-bound on the connectivity of neighboring nodes in $\tilde{G}_L$: 
\begin{lemma}
\label{lemma:graph:homogination2}
When $G_L$ is drawn at random, the probability that an edge connects two arbitrary nodes in $\tilde{G}_L$ is at most $(\lambda \ball(d))^2 p_-$.
\end{lemma}
\begin{proof}
Recall that a pair of nodes from $\tilde{G}_L$, say \((u,v)\), are connected when at least one pair of nodes from corresponding partitions in $G_L$ are connected. Each $d$-neighbourhood in the meaning graph has at most \(\ball(d)\) nodes. It implies that each partition in  $\tilde{G}_L$ has at most \(\lambda \ball(d)\) nodes. Therefore, between each pair of partitions, there are at most $(\lambda \ball(d))^2$ possible edges. By union bound, the probability of at least one edge being present between two partitions is at most $(\lambda \ball(d))^2 p_-$.
\end{proof}

Let $v_w, v_{w'}\in V_{\tilde{G}_L}$ be the nodes corresponding to the components containing $w$ and $w'$ respectively. The following lemma establishes a relation between connectivity of $w,w'\in V_{G_L}$ and the connectivity of $v_s, v_{w'}\in V_{\tilde{G}_L}$: 

\begin{lemma}
\label{lemma:graph:homogination}
\( \prob{ w \pathk{\tilde{d}} w' | m \npathk{} m' }\leq \prob{\text{There is a path from $v_s$ to $v_{w'}$ in $\tilde{G}_L$ with length } \tilde{d} }\). 
\end{lemma}

\begin{proof}
Let $L$ and  \(R\) be the events in the left hand side and right hand side respectively.  
Also for a permutation of nodes in \(G_L\), say \(p\), let \(F_p\) denote the event that all the edges of \(p\) are present, i.e., 
\(L=\cup F_p\).  
Similarly, for a permutation of nodes in \(\tilde{G}_L\), say \(q\), let \(H_q\) denote the event that all the edges of \(q\) are present.
Notice that \(F_p\subseteq H_q\) for \(q\subseteq p\), because if all the edges of \(p\) are present the edges of \(q\) will be present. Thus,
\[L=\bigcup_p F_p\subseteq \bigcup_p H_{p\cap E_{\tilde{G}_L}} =\bigcup_q H_q = R.\]
This implies that \(\prob{L}\leq \prob{R}\).
\end{proof}

{
\begin{lemma}[Upper bound]
\label{lem:UBthm1}
    If $(\lambda\ball(d))^2 p_- \leq \frac{1}{2 e  n}$, then 
    $\prob{ w \pathk{\leq \tilde{d}} w' \mid m \npathk{} m' }
      \leq 2en (\lambda\ball(d))^2 p_-.$
\end{lemma}

\begin{proof}

To identify the upper bound on $\prob{ w \pathk{\leq \tilde{d}} w' | m \npathk{} m' } $, recall the definition of \(\tilde{G}_L\), given an instance of $G_L$ 
(as outlined in Lemma~\ref{lemma:graph:homogination2} and Lemma~\ref{lemma:graph:homogination}, for 
$\tilde{p} = (\lambda \ball(d))^2 p_-$).
Lemma~\ref{lemma:graph:homogination} relates the connectivity of $w$ and $w'$ to a connectivity event in \(\tilde{G}_L\), i.e., \( \prob{ w \pathk{\leq \tilde{d}} w' \mid m \npathk{} m' } \leq \prob{\text{there is a path from $v_s$ to $v_{w'}$ in $\tilde{G}_L$ with length } \tilde{d} }\), where $v_s, v_{w'}\in V_{\tilde{G}_L}$ are the nodes corresponding to the components containing $w$ and $w'$ respectively. Equivalently, in the following, we prove that the event $\text{dist}(v_s,v_{w'})\leq \tilde{d}$ happens with a small probability:

\[
\prob{ w \pathk{\leq \tilde{d}} w' } = 
\prob{ \bigvee_{\ell=1, \cdots, \tilde{d}} w \pathk{\ell} w' }
\leq \sum_{\ell\leq \tilde{d}} { n \choose \ell} \tilde{p}^\ell
\leq \sum_{\ell\leq \tilde{d}} (\frac{e n}{\ell})^\ell \tilde{p}^\ell
\]

\[
\leq \sum_{\ell\leq \tilde{d}} ({e n})^\ell \tilde{p}^\ell
\leq 
{e n\tilde{p}}\frac{
        ({e n\tilde{p}})^{\tilde{d}}-1
    }{
        {e n\tilde{p}}-1
    }
\leq \frac{e n\tilde{p}}{1-e n\tilde{p}}    
\leq 2 e n\tilde{p}. 
\]
where the final inequality uses the assumption that $\tilde{p}\leq \frac{1}{2 e n}$.
\end{proof}

}

Armed with the bounds in Lemma~\ref{lem:LBthm1} and Lemma~\ref{lem:UBthm1}, we are ready to provide the main proof:
\begin{proof}[Proof of 
Theorem~\ref{thm:reasoning:possiblity:detailed}]
Recall that the algorithm checks for connectivity between two given nodes $w$ and $w'$, i.e., 
$w \pathk{\leq \tilde{d}} w'$. With this observation, we aim to infer whether the two nodes in the meaning graph are connected ($m \pathk{\leq d} m'$) or not ($ m \npathk{} m'$). 
We prove the theorem by using lower and upper bound for these two probabilities, respectively:
{ 
\begin{align*}
\gamma 
  & = \prob{  w \pathk{\leq \tilde{d}} w' | m \pathk{d} m' } - \prob{ w \pathk{\leq \tilde{d}} w' | m \npathk{} m' } \\
  & \geq LB\left(\prob{  w \pathk{\leq \tilde{d}} w' | m \pathk{d} m' }\right) - UB\left(\prob{ w \pathk{\leq \tilde{d}} w' | m \npathk{} m' }\right) \\ 
  & \geq \left(1 - 2 e^3 \varepsilon_+ ^{\lambda/2}\right)^{d+1} \cdot  \left(1 - (1-p_+)^{\lambda^2}\right)^{d}- 2en (\lambda\ball(d))^2 p_-. 
\end{align*}
%
where the last two terms of the above inequality are based on the results of Lemma~\ref{lem:LBthm1} and Lemma~\ref{lem:UBthm1}, with the assumption for the latter that $(\lambda\ball(d))^2 p_- \leq \frac{1}{2 e  n}$. To write this result in its general form we have to replace $p_+$ and $p_-$, with $p_+ \oplus \varepsilon_-$ and  $p_- \oplus \varepsilon_-$, respectively (see Remark~\ref{remark:folding}). 
}
\end{proof}

\clearpage

\section{Proofs: Limitations of Connectivity Reasoning}
\label{appendix:proof:connectivity:reasoning}
We provide the necessary lemmas and intuitions before proving the main theorem. 

A random graph is an instance sampled from a distribution over graphs. 
In the $\renymodel$ Erd\H{o}s-Renyi model, a graph is constructed in the following way:  
Each edge is included in the graph with probability $p$, independent of other edges. 
In such graphs, on average, the length of the path connecting any node-pair is short (logarithmic in the number of nodes). 
\begin{lemma}[Diameter of a random graph, Corollary 1 of \cite{chung2002average}]\label{lemma:diameter}
If $n\cdot p = c > 1$ for some constant $c$, then almost-surely the diameter of $\renymodel$ is $\Theta(\log n)$. 
\end{lemma}

We use the above lemma to prove Theorem~\ref{thm:multi-hop:impossiblity}. 
Note that the overall noise probably (i.e., $p$ in Lemma~\ref{lemma:diameter}) in our framework is $p_- \oplus \varepsilon_- $.  

\begin{proof}[Proof of Theorem~\ref{thm:multi-hop:impossiblity}]
Note that the $|V_{G_L}|=\lambda\cdot n$.
By Lemma \ref{lemma:diameter}, the linguistic graph has diameter $\Theta(\log \lambda n)$.
This means that for any pair of nodes $w,w'\in V_{G_L}$, we have $w\pathk{\Theta(\log \lambda n)}w'$.
Since $\tilde{d}\geq\lambda d\in \Omega (\log \lambda n) $, the multi-hop reasoning algorithm finds a path between $w$ and $w'$ in linguistic graph and returns $\mathsf{connected}$ regardless of the connectivity of $m$ and $m'$.
\end{proof}

\clearpage

\section{Proofs: Limitations of General Reasoning}
\label{appendix:proof:general:reasoning}
The proof of the general limitations theorem follows after introducing necessary notation and lemmas. This section is structured to first cover key definitions and two main lemmas that lead to the theorem, after which proofs of auxiliary lemmas and results are included to complete the formal argument.

\subsection{Main Argument}

\begin{figure}[h]
    \centering
    \includegraphics[scale=0.39, trim=0cm 0.3cm 0cm 0cm]{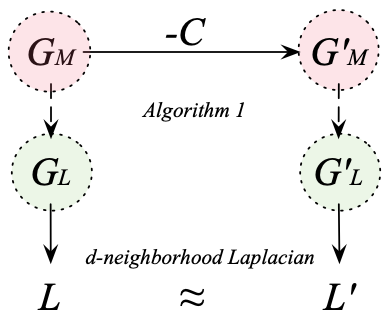}
    \caption{
    The construction considered in Definition~\ref{cut:construction}. 
    The node-pair $m$-$m'$ is connected with distance $d$ in $G_M$, and disconnected in $G_M'$, after dropping the edges of a cut $C$. For each linguistic graph, we consider it ``local'' Laplacian. 
    }
    \label{fig:laplacians}
\end{figure}

Consider a meaning graph $G_M$ in which two nodes $m$ and $m'$ are connected. We drop edges in a min-cut $C$ to make the two nodes disconnected and obtain $G_M'$ (Figure~\ref{fig:laplacians}). 

\begin{definition}
\label{cut:construction}
Define a \emph{pair} of meaning graphs $G$ and $G'$, both with $n$ nodes and satisfying ball assumption $\ball(d)$, with three properties: (1) $m \pathk{d} m'$ in $G$, (2) $m \npathk{} m'$ in $G'$, (3) $E_{G'} \subset E_G$, (4) $C = E_G \setminus E_{G'}$ is an $(m, m')$ min-cut of $G$. 
\end{definition}


\begin{definition}
\label{def:distribution}
Define a distribution $\mathcal{G}$ over pairs of possible meaning graphs $G, G'$ and pairs of nodes $m,m'$ which satisfies the requirements of Definition \ref{cut:construction}. Formally, \(\mathcal{G}\) is a uniform distribution over the following set:
\vspace{-0.05cm}
{
\small
$$ 
\{(G,G',m,m') \mid G,G',m,m' \text{satisfy Definition \ref{cut:construction}}\}.
$$
}
\end{definition}
\newcommand{\dimm}{|\mathcal{U}|}

As two linguistic graphs, we sample $G_L$ and $G'_L$, as denoted in Figure~\ref{fig:laplacians}. 
In the sampling of $G_L$ and $G'_L$, all the edges share the randomization, except for the ones that correspond to $C$ (i.e., the difference between the $G_M$ and $G'_M$).  
Let $\mathcal{U}$ be the union of the nodes involved in $\tilde{d}$-neighborhood of $w, w'$, in $G_L$ and $G'_L$. 
Define $L, L'$ to be the Laplacian matrices corresponding to the nodes of $\mathcal{U}$. As $n$ grows, the two Laplacians become less distinguishable whenever \changed{$p_- \oplus  \varepsilon_-$} and $d$ are large enough:

\begin{lemma}
\label{lem:normLaplacianClose}
Let $c > 0$ be a constant and $p_-, \lambda$ be parameters of the sampling process in Algorithm~\ref{al:sampling} on a pair of meaning graphs $G$ and $G'$ on $n$ nodes constructed according to Definition~\ref{cut:construction}. Let $d \in \mathbb{N}, \tilde{d} \geq \lambda d,$ and $L, L'$ be the Laplacian matrices for the $\tilde{d}$-neighborhoods of the corresponding nodes in the sampled linguistic graphs $\gl$ and $\gl'$. If
$\changed{ p_- \oplus \varepsilon_-} \geq \frac{c \log n}{n}$ and 
$d > \log n,$
then, with a high probability, the two Laplacians are close:
$$
\vspace{-0.05cm}
\norm{\tilde{L}-\tilde{L'}} \leq \frac{ \sqrt{2\lambda}\, \ball(1)}{ \sqrt{n \log (n \lambda)} } 
\vspace{-0.05cm}
$$
\end{lemma}

\begin{proof}[Proof of Lemma~\ref{lem:normLaplacianClose}]
We start by proving an upper bound on $\tilde{L}-\tilde{L'}$ in matrix inequality notation. Similar upper-bound holds for  $\tilde{L'}-\tilde{L}$ which concludes the theorem.
\begin{align*}
\tilde{L}-\tilde{L'} &= \frac{L}{\norm{{L}}}-\frac{L'}{\norm{L'}} \\ 
& \preceq \frac{L}{\norm{L}}-\frac{L'}{\norm{L-L'}+\norm{L}} \\ 
& =\frac{L\cdot  \norm{L-L'}}{\norm{L}^2}+\frac{L-L'}{\norm{L}}\\ 
& \preceq \frac{ \sqrt{\lambda} \ball(1)}{ \sqrt{2n \log (n \lambda)} }I+\frac{ \sqrt{\lambda} \ball(1)}{ \sqrt{2n \log (n \lambda)} }I.
\end{align*}
The last inequality is due to Lemma \ref{lemma:impossiblity:general:reasoning}.
By symmetry the same upper-bound holds for \(\tilde{L}'-\tilde{L}\preceq
2\frac{ \sqrt{\lambda} \ball(1)}{ \sqrt{2n \log (n \lambda)} }I\). This means that \(\norm{\tilde{L}-\tilde{L}'}\leq \frac{ 2\sqrt{\lambda} \ball(1)}{ \sqrt{2n \log (n \lambda)} }\). 
\end{proof}

This can be used to show 
that, for such large enough $p_-$ and $d$, the two linguistic graphs, $G_L$ and $G'_L$ sampled as above, are indistinguishable by any 
function operating over a $\lambda d$-neighborhood of $w, w'$ in \gl, with a high probability.

A reasoning function can be thought of a mapping defined on normalized Laplacians, 
since they encode all the information in a graph.  
For a reasoning function $f$ with limited precision, the input space can be partitioned into regions where the function is constant; and for large enough values of $n$ both $\tilde{L}, \tilde{L}'$ (with a high probability) fall into regions where $f$ is constant.

Note that a reasoning algorithm is oblivious to the the details of $C$, i.e. it does not know where $C$ is, or where it has to look for the changes. Therefore a realistic algorithm ought to use the neighborhood information collectively.

In the next lemma, we define a function $f$ to characterize the reasoning function, which uses Laplacian information and maps it to binary decisions. We then prove that for any such functions, there are regimes that the function won't be able to distinguish $\tilde{L}$ and $\tilde{L}'$: 

\begin{lemma}
\label{lemma:improbable:recovery}
Let meaning and linguistic graphs be constructed under the conditions of Lemma~\ref{lem:normLaplacianClose}. Let $\beta > 0$ and $f: \mathbb{R}^{\dimm \times \dimm} \rightarrow \{0, 1\}$ be the indicator function of an open set.
Then there exists $n_0 \in \mathbb{N}$ such that for all $n \geq n_0$: 
$$
\mathbb{P}_{ \substack{ (G,G',m,m') \sim \mathcal{G} \\ G_L \leftarrow \alg{G}, \\G'_L \leftarrow \alg{G'}  } }\left[ f(\tilde{L}) = f(\tilde{L}') \right] \geq 1 - \beta.
$$
\end{lemma}

\begin{proof}[Proof of Lemma~\ref{lemma:improbable:recovery}]

Note that $f$ maps a high dimensional continuous space to a discrete space. To simplify the argument about $f$,
we decompose it to two functions: a continuous function $g$ mapping matrices to $(0,1)$ and a threshold function $H$ (e.g. $0.5 + 0.5 \text{sgn}(.)$) which maps to one if $g$ is higher than a threshold and to zero otherwise. Without loss of generality we also normalize $g$ such that the gradient is less than one. Formally, 
$$
f = H \circ g, \text{ where } g: \mathbb{R}^{\dimm \times \dimm } \rightarrow (0, 1), \norm{ \nabla g \Bigr|_{\tilde{L}}}\leq 1.
$$
Lemma~\ref{lemma:compositon} gives a proof of existence for such decompositon, which depends on having open or closed pre-images. 

One can find a differentiable and Lipschitz function $g$ such that 
it intersects with the threshold specified by $H$, in the borders where $f$ changes values.  

With $g$ being Lipschitz, one can upper-bound the variations on the continuous function: 
$$
\norm{g(\tilde{L}) - g(\tilde{L'})} \leq M \norm{\tilde{L} - \tilde{L}'}. 
$$
According to Lemma~\ref{lem:normLaplacianClose}, $\norm{\tilde{L} - \tilde{L}'}$ is upper-bounded by a decreasing function in $n$. 

For uniform choices $(G,G',m,m') \sim \mathcal{G}$ (Definition~\ref{def:distribution}) the Laplacian pairs $(\tilde{L}, \tilde{L}')$ are randomly distributed in a high-dimensional space, and for big enough $n$, there are enough portion of the $(\tilde{L}, \tilde{L}')$  (to satisfy $1-\beta$ probability) that appear in the same side of the hyper-plane corresponding to the threshold function (i.e. $f(\tilde{L}) = f(\tilde{L}')$). 
\end{proof}

 {
  \begin{proof}[Proof of Theorem~\ref{thm:impossiblity:general:reasoning}]
    
    By substituting $\gamma=\beta$, and $X_L=f$, the theorem reduces to the statement of Lemma~\ref{lemma:improbable:recovery}.

\end{proof}
 
}

\subsection{Auxiliary Lemmas and Proofs}

In the following lemma, we show that the spectral differences between the two linguistic graphs in the locality of the target nodes are small. 
For ease of exposition, we define an intermediate notation, for a normalized version of the Laplacians: $\tilde{L} = L / \| L\|_2$ and $\tilde{L}' = L' / \| L'\|_2$.

\begin{lemma}
\label{lemma:cut-norm}
The norm-2 of the Laplacian matrix corresponding to the nodes participating in a cut, can be upper-bounded by the number of the edges participating in the cut (with a constant factor). 
\end{lemma}

\begin{proof}[Proof of Lemma~\ref{lemma:cut-norm}]
Using the definition of the Laplacian: 
$$
\|L_C\|_2 \leq \| A - D \|_2 \leq   \| A \|_2 + \| D \|_2
$$
where $A$ is the adjacency matrix and $D$ is a diagonal matrix with degrees on the diagonal. We bound the norms of the matrices based on size of the cut (i.e., number of the edges in the cut). For the adjacency matrix we use the Frobenius norm:  
$$
\| A \|_2 \leq  \| A \|_F = \sqrt{\sum_{ij} a_{ij}} = 2 \cdot |C|
$$
where $|C|$ denotes the number of edges in $C$.
To bound the matrix of degrees, we use the fact that norm-2 is equivalent to the biggest eigenvalue, which is the biggest diagonal element in a diagonal matrix: 
$$
\| D \|_2 = \sigma_{\max}(D) = \max_i deg(i) \leq |C|
$$
With this we have shown that: $\|L_C\|_2 \leq 3|C|$. 
\end{proof}

For sufficiently large values of $p$, $\renymodel$ is a connected graph, with a high probability. More formally: 
\begin{lemma}[Connectivity of random graphs]
\label{lemma:er:connectedness}
In a random graph $\renymodel$, for any $p$ bigger than ${{\tfrac {(1+\varepsilon )\ln n}{n}}}$, the graph will almost surely be connected.
\end{lemma}
The proof can be found in \cite{erdos1960evolution}. 

\begin{lemma}[Norm of the adjacency matrix in a random graph]
\label{lemma:adj:norm}
For a random graph $\renymodel$, let $L$ be the adjacency matrix of the graph. 
For any $\varepsilon > 0$: 
$$
\lim_{n \rightarrow +\infty} \mathbb{P} \left( \left| \|L\|_2   - \sqrt{2n\log n} \right| > \varepsilon \right)  \rightarrow 0
$$
\end{lemma}
\begin{proof}[Proof of Lemma~\ref{lemma:adj:norm}]
From Theorem 1 of~\cite{ding2010spectral} we know that: 
$$
\frac{ \sigma_{\max}{(L)} }{ \sqrt{n \log n}}  \overset{P}{\rightarrow} \sqrt{2}
$$
where $\overset{P}{\rightarrow}$ denote \emph{convergence in probability}. 
And also notice that norm-2 of a matrix is basically the size of its biggest eigenvalue, which concludes our proof. 
\end{proof}

\begin{lemma}
\label{lemma:impossiblity:general:reasoning}
For any pair of meaning-graphs $G$ and $G'$ constructed according to Definition~\ref{cut:construction}, and,  
\begin{itemize}
    \item $d > \log n$, 
    \item $p_- \oplus \varepsilon_- \geq c \log n \big/ n$ for some constant $c$, 
    \item $\tilde{d}\geq \lambda d$, 
\end{itemize}
with $L$ and $L'$ being the Laplacian matrices corresponding to the $\tilde{d}$-neighborhoods of the corresponding nodes in the surface-graph; we have: 
$$
\frac{\|L-L'\|_2}{\|L\|_2} \leq \frac{ \sqrt{\lambda} \ball(1)}{ \sqrt{2n \log (n \lambda)} }, 
$$
with a high-probability.
\end{lemma}

\begin{proof}[Proof of Lemma~\ref{lemma:impossiblity:general:reasoning}]
In order to simplify the exposition, w.l.o.g. assume that $\varepsilon_- =0$ (see Remark~\ref{remark:folding}). 
Our goal is to find an upper-bound to the fraction $\frac{\|L-L'\|_2}{\|L\|_2}$. 
Note that the Laplacians contain only the local information, i.e., $\tilde{d}-$neighborhood. 
First we prove an upper bound on the nominator. 
By eliminating an edge in a meaning-graph, the probability of edge appearance in the linguistic graph changes from $p_+$ to $p_-$. 
The effective result of removing edges in $C$ would appear as i.i.d. $\bern{p_+-p_-}$. 
Since by definition, $\ball(1)$ is an upper bound on the degree of meaning nodes, the size of minimum cut should also be upper bounded by \ball(1).
Therefore, the maximum size of the min-cut $C$ separating two nodes $m \pathk{d} m'$ is at most $\ball(1)$. 
To account for vertex replication in linguistic graph, the effect of cut would appear on at most $\lambda \ball(1)$ edges in the linguistic graph. Therefore, we have$\|L-L'\|_2\leq \lambda \ball(1)$ using Lemma~\ref{lemma:cut-norm}.


As for the denominator, the size of the matrix $L$ is the same as the size of $\tilde{d}$-neighborhood in the linguistic graph. 
We show that if $\tilde{d} > \log (\lambda n)$  the neighborhood almost-surely covers the whole graph. 
While the growth in the size of the $\tilde{d}$-neighborhood is a function of both $p_+$ and $p_-$, to keep the analysis simple, we underestimate the neighborhood size by replacing $p_+$ with $p_-$, i.e., the size of the $\tilde{d}$-neighborhood is lower-bounded by the size of a $\tilde{d}$-neighborhood in $\mathsf{G}(\lambda \cdot n, p_-)$. 

By Lemma~\ref{lemma:er:connectedness} the diameters of the linguistic graphs $G_L$ and $G_L'$ are both $\Theta(\log (\lambda n))$.  
Since $\tilde{d} \in \Omega(\log (\lambda n))$, $\tilde{d}$-neighborhood covers the whole graph for both $G_L$ and $G_L'$.

Next, we use Lemma~\ref{lemma:adj:norm} to state that $\|L\|_2$ converges to $\sqrt{2 \lambda n \log (\lambda n)}$, in probability. 

Combining numerator and denominator, we conclude that the fraction, for sufficiently large $n$, is upper-bounded by: 
$\frac{
\lambda\ball(1)
}{
\sqrt{2 \lambda n \log (\lambda n)}
},$
which can get arbitrarily small, for a big-enough choice of $n$. 
\end{proof}






\begin{lemma}
\label{lemma:compositon}
Suppose $f$ is an indicator function on an open set\footnote{\url{https://en.wikipedia.org/wiki/Indicator_function}}, 
it is always possible to write it as composition of two functions:
\begin{itemize}
   \item A continuous and Lipschitz function: $g: \mathbb{R}^{d} \rightarrow (0, 1), $
   \item A thresholding function: $H(x)= \mathbf{1} \{ x > 0.5 \}.$
\end{itemize}
such that: $ \forall x \in \mathbb{R}^d  :  \; \; \;  f(x) = h(g(x))$. \end{lemma}
\begin{proof}[Proof of Lemma~\ref{lemma:compositon}]
Without loss of generality, we assume that the threshold function is defined as $H(x) = \mathbf{1} \{ x > 0.5 \}$. One can verify that a similar proof follows for $H(x) = \mathbf{1} \{ x \geq 0.5 \}$. 
We use notation $f^{-1}(A)$ the set of pre-images of a function $f$, for the set of outputs $A$. 

First let's study the collection of inputs that result in output of $1$ in $f$ function. 
Since $f=h\circ g$, then $f^{-1}(\{1\})=g^{-1}(h^{-1}(\{1\}))=g^{-1}((0.5,1))$ and 
$f^{-1}(\{0\})=g^{-1}(h^{-1}(\{0\}))=g^{-1}((0,0.5))$.
Define $C_0$ and $C_1$, such that $C_i \triangleq f^{-1}(\{i\})$; note that  
since $g$ is continuous and $(0.5,1)$ is open $C_1$ is an open set (hence $C_1$ is closed). 
Let
$d:\mathbb R^n\to \mathbb R$ be defined by, 
$$
d(x) \triangleq \text{dist}(x,C_0)=\inf_{c\in C_0}\|x-c\|. 
$$
Since $C_0$ is closed, it follows $d(x)=0$ if and only if $x\in C_0$. Therefore, letting $$g(x)=\frac12 + \frac12\cdot \frac{d(x)}{1+d(x)},$$
then $g(x)=\frac12$ when $x\in C_0$, while $g(x)>\frac12$ when $x\not\in C_0$. This means that letting $h(x)=1$ when $x>\frac12$ and $h(x)=0$ when $x\le \frac12$, then $f=h\circ g$.
One can also verify that this construction is $1/2$-Lipschitz; this follows because $d(x)$ is $1$-Lipschitz, which can be proved using the triangle inequality

Hence the necessary condition to have such decomposition is $f^{-1}(\{1\})$ and $f^{-1}(\{0\})$ be open or closed. 
\end{proof}

\clearpage

\section{Empirical investigation: further details}
\label{appendix:empirical:details}

\begin{figure*}[h]
    \centering
    \includegraphics[scale=0.38]{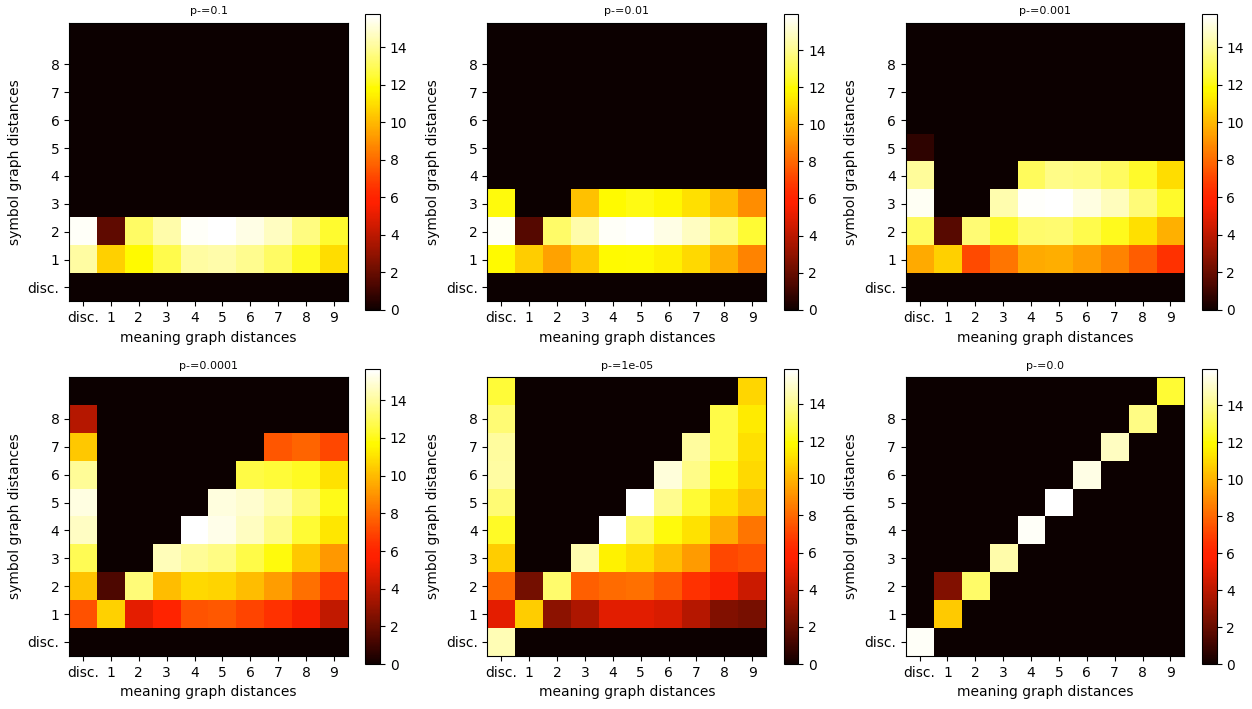}
    \caption{With varied values for $p_-$ a heat map representation of the distribution of the average distances of node-pairs in linguistic graph based on the distances of their corresponding meaning nodes is presented. }
    \label{fig:exp:d:vs:d}
\end{figure*}

To evaluate the impact of the other noise parameters in the sampling process, we compare the average distances between nodes in the linguistic graph for a given distance between the meaning graph nodes. In the Figure~\ref{fig:exp:d:vs:d}, we plot these graphs for decreasing values of $p_-$ (from top left to bottom right). With high $p_-$ (top left subplot), nodes in the linguistic graph at distances lower than two, regardless of the distance of their corresponding node-pair in the meaning graph. As a result, any reasoning algorithm that relies on connectivity can not distinguish linguistic nodes that are connected in the meaning space from those that are not. As the $p_-$ is set to lower values (i.e. noise reduces), the distribution of distances get wider, and correlation of distance between the two graphs increases. In the bottom middle subplot, when $p_-$ has a very low value, we observe a significant correlation that can be reliably utilized by a reasoning algorithm.

